\documentclass{article}

\PassOptionsToPackage{numbers, compress}{natbib}

\usepackage[preprint]{my_preprint}

\usepackage[utf8]{inputenc} 
\usepackage[T1]{fontenc}    
\usepackage{hyperref}       
\usepackage{url}            
\usepackage{booktabs}       
\usepackage{amsfonts}       
\usepackage{nicefrac}       
\usepackage{microtype}      

\usepackage{amsmath}
\usepackage{amssymb}
\usepackage{adjustbox}

\usepackage{booktabs}       
\usepackage{multirow}       
\usepackage{xcolor}         
\usepackage{colortbl}
\usepackage{graphicx}       
\usepackage{tabularx}

\usepackage{listings}
\usepackage{setspace}
\usepackage{algorithmic}
\usepackage[ruled,linesnumbered]{algorithm2e}
\usepackage{caption}
\usepackage[compact]{titlesec}
\usepackage{wrapfig}
\usepackage{makecell}

\newcommand{\systemname}{NeuralDiffuser}

\title{\systemname: 
Neuroscience-inspired Diffusion Guidance for fMRI Visual Reconstruction}

\author{%
  Haoyu Li\\
  Institute of Artificial Intelligence and Robotics\\
  Xi'an Jiaotong University\\
  \texttt{lihy98@stu.xjtu.edu.cn} \\
  \And
  Hao Wu\thanks{Corresponding authors.}\\
  School of Electrical Engineering\\
  Xi'an University of Technology\\
  \texttt{wuhaoacm@163.com} \\
  \And
  Badong Chen\footnotemark[1]\\
  Institute of Artificial Intelligence and Robotics\\
  Xi'an Jiaotong University\\
  \texttt{chenbd@mail.xjtu.edu.cn} \\
}

\begin{document}

\maketitle

\begin{abstract}
  Reconstructing visual stimuli from functional Magnetic Resonance Imaging (fMRI) enables fine-grained retrieval of brain activity. However, the accurate reconstruction of diverse details, including structure, background, texture, color, and more, remains challenging. The stable diffusion models inevitably result in the variability of reconstructed images, even under identical conditions. To address this challenge, we first uncover the neuroscientific perspective of diffusion methods, which primarily involve top-down creation using pre-trained knowledge from extensive image datasets, but tend to lack detail-driven bottom-up perception, leading to a loss of faithful details. In this paper, we propose NeuralDiffuser, which incorporates primary visual feature guidance to provide detailed cues in the form of gradients. This extension of the bottom-up process for diffusion models achieves both semantic coherence and detail fidelity when reconstructing visual stimuli. Furthermore, we have developed a novel guidance strategy for reconstruction tasks that ensures the consistency of repeated outputs with original images rather than with various outputs. Extensive experimental results on the Natural Senses Dataset (NSD) qualitatively and quantitatively demonstrate the advancement of NeuralDiffuser by comparing it against baseline and state-of-the-art methods horizontally, as well as conducting longitudinal ablation studies. Code can be available on \href{https://github.com/HaoyyLi/NeuralDiffuser}{https://github.com/HaoyyLi/NeuralDiffuser}.
\end{abstract}

\section{Introduction}
\label{Introduction}
Decoding visual stimuli from brain responses, such as functional magnetic resonance imaging (fMRI), holds significant promise in the field of neuroscience~\cite{wu2020encoding}. 
Recently, researchers have demonstrated a keen interest in a compelling decoding task that aims to reconstruct visual stimuli perceived by subjects, thereby providing a fine-grained retrieval of the intricate connections between biological brain activity and cognitive computation. 
However, this endeavor remains challenging due to the difficulty of learning a mapping from fMRI to images using neural networks with suboptimal representation capabilities and insufficient fMRI image pairs. Numerous pixel-to-pixel methods produce ambiguous images that lack semantic coherence~\cite{st2018generative, seeliger2018generative, shen2019end, lin2019dcnn, ren2021reconstructing}.

\begin{figure}[!t]
    \begin{center}
        \centerline{\includegraphics[width=0.8\linewidth]{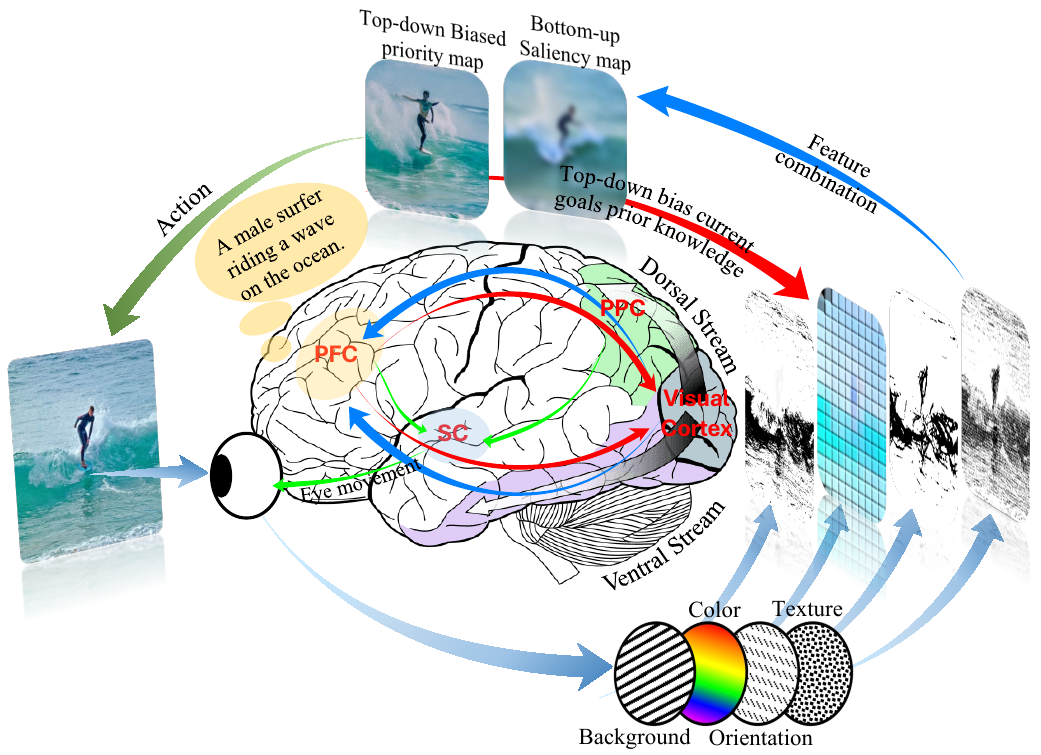}}
    \end{center}
    \caption{Schematic diagram of bottom-up and top-down processes in neuroscience. The perception of the visual scene is shaped by the reciprocal interaction of bottom-up perception, driven by visual cues from the retina ({blue flows}), and top-down creation, which incorporates prior knowledge and experience ({red flows}).}
    \label{fig:neuralscience}
\end{figure}

\begin{figure*}[!t]
    \begin{center}
        \centerline{\includegraphics[width=\linewidth]{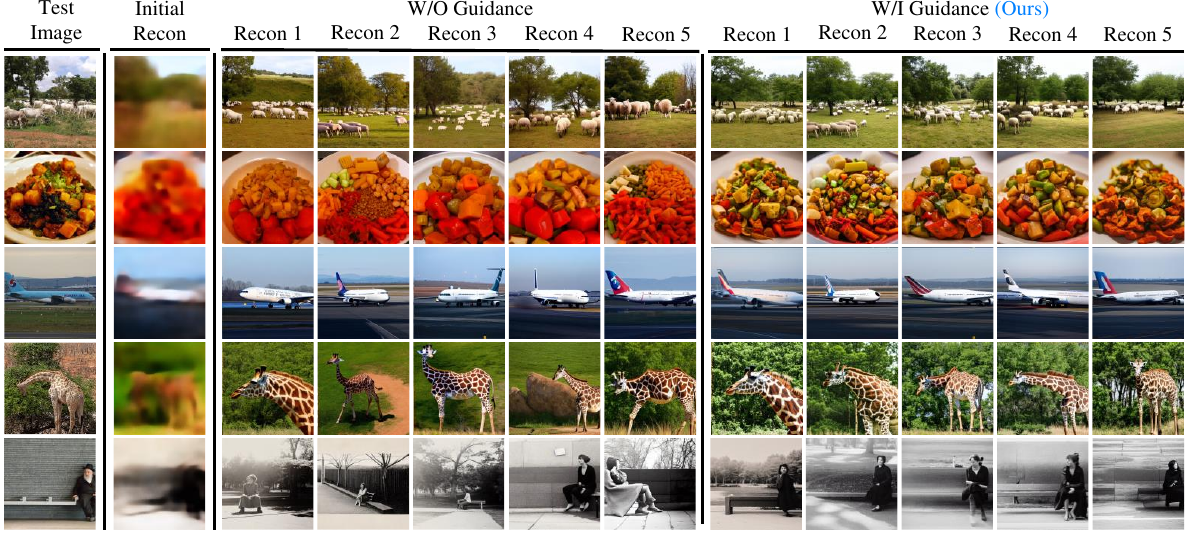}}
    \end{center}
    \caption{Reconstructed examples of diffusion-based (left) and \systemname\ (right). Diffusion-based results are initialized by blurry initial images but lack detail cues, thereby tending to produce different 5 images. In contrast, \systemname\ proposes primarily visual feature guidance, thereby obtaining faithful details and consistent 5 original images.}
    \label{fig:fig1}
\end{figure*}

Nowadays, the accelerated progression of generative models, including stable diffusion models~\cite{rombach2022high}, has engendered a fertile landscape for achieving substantial advancements in the reconstruction of visual stimuli. The pre-trained Latent Diffusion Model (LDM) capitalizes on prior knowledge and experience learned from extensive image datasets to generate high-quality natural images utilizing two inputs: the low-level initial latent embedding and the high-level semantic condition. Several studies have successfully reconstructed visual stimuli by aligning the fMRI data with the two LDM embeddings using ridge regression~\cite{takagi2023high, ozcelik2023natural, lu2023minddiffuser}. In MindEye~\cite{scotti2023reconstructing}, diffusion priors in DALL-E2~\cite{ramesh2022hierarchical} are introduced to address the modality gap between fMRI and the latent space of LDM, leading to promising results.

However, two significant challenges persist: Firstly, for model training, fMRI image pairs are inadequate for establishing a complex mapping between fMRI and LDM embeddings. Although the large-scale fMRI dataset, such as the Natural Senses Dataset (NSD)~\cite{allen2022massive}, has been made public, fMRI image pairs are still limited because individual-specific skull sizes lead to a different number of voxels that cannot be fed into the model simultaneously. Secondly, the reconstruction of the details of visual stimuli still requires further improvement. While stable-diffusion models can generate images with similar semantics to the original images, aligning their details remains a challenge. To address this,  MindDiffuser~\cite{lu2023minddiffuser} and MindEye-BOI~\cite{kneeland2023brain} have proposed iterative diffusion models to optimize the two input embeddings, though this process is notably time-consuming. In contrast, Takagi~\cite{takagi2023improving} and DREAM~\cite{xia2024dream} propose that incorporating depth and color features enhances detail fidelity. However, decoding these features from fMRI presents significant challenges.

The perspective on diffusion-based fMRI reconstruction presented herein is grounded in neuroscientific theory. Visual cognition involves both a bottom-up process, which involves the passive perception of visual cues from the thalamus, and a top-down process, which involves the active imagination by the conscious mind, as illustrated in Fig.\ref{fig:neuralscience}. 
The top-down and bottom-up processes are intertwined in the human visual system. The process of generating images within the two input embeddings via the stable diffusion embodies the top-down process, where pre-training knowledge and experience on large-scale datasets enhance the semantic coherence of images. In terms of bottom-up processing, the low-level initial latent embedding provides a starting point, while there is no interweaving with top-down processes during diffusion. Furthermore, decoding the initial latent embedding from fMRI is often inaccurate (Fig.\ref{fig:fmri_decoding}), and the blurred initial latent may misdirect the reconstructed images, as demonstrated in Fig.~\ref{fig:fig1} (left: w/o guidance). To make matters worse, the stable diffusion model is a generative model that is inherently diverse, generating different images even under the same conditions.

In this paper, we propose a diffusion guidance for visual reconstruction, termed \systemname, which adds detailed cues to diffusion models, achieving both semantic coherence and detail fidelity, resulting in consistent results with the original images. Specifically, \systemname\ follows subject-specific voxel encoders to map voxels to a subject-shared space and then aligns fMRI embeddings with the latent space of LDM by fMRI decoders. Decoders are pre-trained across subjects and fine-tuned on the target subject, leveraging all subjects' fMRI-image pairs to enhance fMRI decoding performance. Additionally, a primary visual feature guidance is proposed, incorporating detailed visual cues to control the reconstruction process. Furthermore, a novel guidance strategy for the reconstruction task is developed, ensuring detail fidelity to the original image and consistency across repeated outputs, as illustrated in Fig.~\ref{fig:fig1} (right: w/i guidance).

From a neuroscience perspective, \systemname\ proposes primary visual feature guidance, which opens up a feedforward bottom-up process to provide detailed cues and combines it with a knowledge-driven top-down process. From a computational viewpoint, \systemname\ incorporates primary visual features that are well-decoded to compensate for suboptimal decoding performance on blurry initial latents and to enhance the fidelity of the reconstructed details.

The proposed \systemname\ offers three distinct advantages. First, it employs primary visual feature guidance to control the diffusion process, thereby achieving impressive semantic coherence, detail fidelity, and repeat consistency. Second, it develops a novel guidance strategy to obtain consistent results with the original images, which can be extended to other reconstruction tasks. Third, the proposed method is independent of stable diffusion and fMRI embeddings, enabling its adoption in any existing diffusion-based reconstruction approach.

Our contributions can be summarized as follows:
\begin{itemize} 
\item Constructive insights are offered into diffusion models from the perspective of neuroscience theory, with a particular focus on top-down and bottom-up processes. The reasons for the unfaithful details in diffusion-based fMRI reconstruction methods are elucidated.
\item \systemname\ is proposed, which introduces primary visual feature guidance to control the diffusion process and provides comprehensive instructions for fMRI reconstruction. For reconstruction tasks, a novel guidance strategy is developed to enhance detail fidelity and repeat consistency. 
\item Extensive experimental results demonstrate the advancements of \systemname\ qualitatively and quantitatively by comparing horizontally (baseline and state-of-the-art methods) and longitudinally (ablation studies).
\end{itemize}

\section{Related Work}
\subsection{fMRI Visual Reconstruction}
\begin{figure}[!t]
    \begin{center}
        \centerline{\includegraphics[width=.9\linewidth]{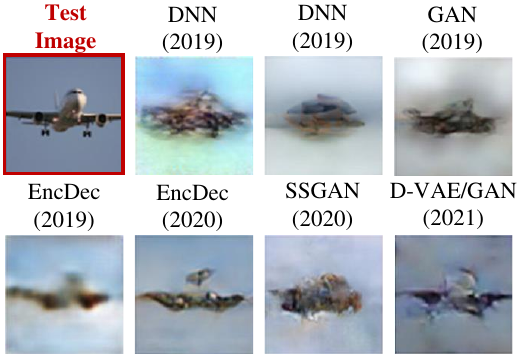}}
    \end{center}
    \caption{{Reconstructed images using VAEs and GANs}}
    \label{fig:gans_vaes}
\end{figure}

\textbf{Pixel-to-Pixel:} 
Numerous previous studies employ generative adversarial networks (GANs)~\cite{goodfellow2014generative}, variational autoencoders (VAEs)~\cite{kingma2013auto}, and self-supervised learning~\cite{beliy2019voxels} to achieve pixel-level reconstruction of visual stimuli~\cite{st2018generative, seeliger2018generative, shen2019end, lin2019dcnn, ren2021reconstructing}. However, pixel-level models are constrained by limited representation within insufficient fMRI image pairs due to the disparity between voxels in the visual cortex and image pixels. Consequently, such methods often yield ambiguous images and lack semantic meaning, as illustrated in Fig.~\ref{fig:gans_vaes}.

\textbf{Pretrained Generative Models:} 
Deep generative models, including IC-GAN~\cite{perarnau2016invertible}, StyleGAN~\cite{karras2020training, karras2020analyzing}, and Stable Diffusion~\cite{rombach2022high}, have been extensively used in visual reconstruction tasks, where MindReader~\cite{lin2022mind} fine-tunes the models to align with the representations of fMRI. 
Alternatively, some methods map fMRI to IC-GAN~\cite{ozcelik2022reconstruction, gu2023decoding} or Stable Diffusion~\cite{takagi2023high, chen2023seeing, lu2023minddiffuser, ozcelik2023natural} using ridge regression. Furthermore, recent studies have demonstrated the alignment of fMRI to diffusion models using a large neural network, which has yielded notable results~\cite{scotti2023reconstructing, scotti2024mindeye2, liu2024see}. MindEye2~\cite{scotti2024mindeye2} and Liu \emph{et al.}~\cite{liu2024see} used subject adapters to align subject-dependent voxels in a shared space, thereby enabling cross-subject transfer learning. \systemname\ is built upon Stable Diffusion, as in the above studies. To ensure detail fidelity, we propose primary visual feature guidance inspired by neuroscience theory and develop a guidance strategy for reconstruction tasks.

\subsection{Diffusion Models}
\textbf{Conditional Image Generation:} 
Diffusion models are provided with conditions (categories or prompts) in the form of classifier~\cite{dhariwal2021diffusion} or classifier-free~\cite{ho2022classifier} guidance. 
Classifier guidance involves the back-propagation of gradients through additional trained classifiers that can process noisy images. Classifier-free guidance, on the other hand, retrains the diffusion model to input semantic embeddings directly, thus eliminating the need for multiple additional classifiers.

\textbf{Guided Image Generation:} 
To control the generated images by visual properties (sketch, depth map, edge map, style, color, etc.), studies have trained control models to guide diffusion models~\cite{zhang2023adding, mou2023t2i, zhao2023uni}. Furthermore, some studies have fine-tuned the UNet to input guidance using concatenation~\cite{ju2023humansd, voynov2023sketch} or cross-attention~\cite{zheng2023layoutdiffusion}. However, these methods require training a control model or fine-tuning the diffusion model. Moreover, decoding the guidance of visual properties directly from fMRI is challenging~\cite{takagi2023improving, xia2024dream}. In contrast, a general guidance method has been proposed to extend the control function~\cite{bansal2023universal}. The objective of our research is to reconstruct the original images, as opposed to generating new images as in the aforementioned studies. Furthermore, the decoding of guidance from fMRI inevitably results in inaccurate control and additional noise.

\section{Preliminary}
\label{Preliminary}
\subsection{Visual Neuroscience}
A comprehensive understanding of how the brain perceives visual stimuli necessitates the elucidation of the dynamic connections, both top-down and bottom-up~\cite{pollen1999neural}, as illustrated in Fig.~\ref{fig:neuralscience}. Bottom-up processes in neuroscience refer to how sensory information is processed commencing from the stimulus itself, traversing sensory pathways, and culminating in the brain. The initial visual processing in the primary visual cortex (V1) involves the detection of edges, colors, and motion, which are critical for building a coherent visual representation. For example, when an object is perceived, the visual system first detects edges and colors before the brain recognizes it as a specific item, like an apple. This is driven by the raw data received from the environment. In contrast, top-down processes involve higher cognitive functions influencing perception. The brain utilizes prior knowledge to filter and interpret sensory information, thereby creating a coherent perceptual experience~\cite{gilbert2007brain}. An illustration of this phenomenon can be observed in the manner in which context can influence our interpretation of ambiguous images, as exemplified by the "Rubin's vase" illusion, wherein perception undergoes a shift between two distinct interpretations based on prior experiences. The bottom-up and top-down processes are known to conjugate synchronously during the processing of visual stimuli by the brain. Low-level cortical areas are adept at processing direct light sensing, while high-level cortical areas have been observed to specialize in semantic processing~\cite{pollen1999neural}. Studies indicate that low-level feature selectivity plays a crucial role in computing high-level semantics, and the statistics of primary visual features are modulated by semantic content~\cite{henderson2023low}. As a specific example, outdoor scenes often exhibit prominent horizontal orientations and high spatial frequencies. These observations suggest that the two aforementioned processes can synergistically contribute to a more comprehensive understanding of various visual stimuli.

\subsection{LDM's Neurological Perspective}
The LDM embeddings are constructed on the CLIP shared space for the high-level semantic condition $c$ and the VAE's latent space for the low-level initial latent $z$. The LDM generates images based on $z$ and $c$. Specifically, $z$ is first diffused forward to Gaussian noise $z_T$. A pre-trained U-Net is then used to estimate the noise at the $t$-th step conditional on $c$ and remove noise from $z_t$. Finally, image results are decoded from $z_0$ using the VQ-VAE decoder. For top-down processes, LDM generates images by sampling from a conditional probability distribution $p(x|c)$. It is posited that $p(x|c)$ represents the knowledge and experience pre-trained on massive images, thus promoting the semantic coherence of fMRI reconstruction. In terms of bottom-up processes, the latent space of VAE is indeed a discrete representation space for images rather than primary visual features. Consequently, decoding the initial latent $z$ directly from fMRI is equivalent to pixel-level reconstruction, which results in blurry images that are inadequate for providing detailed features in a bottom-up context. Furthermore, the VAE latent initializes the reverse diffusion process but does not conjugate with high-level semantic conditions. Consequently, we contend that the core principle of LDM lies in its fidelity to semantics. However, this approach entails a compromise in detail, which may impede the attainment of faithful details and consistent results in reconstruction tasks.

\section{Methods}
\label{Methods}

\begin{figure*}[!t]
    \begin{center}
        \centerline{\includegraphics[width=\linewidth]{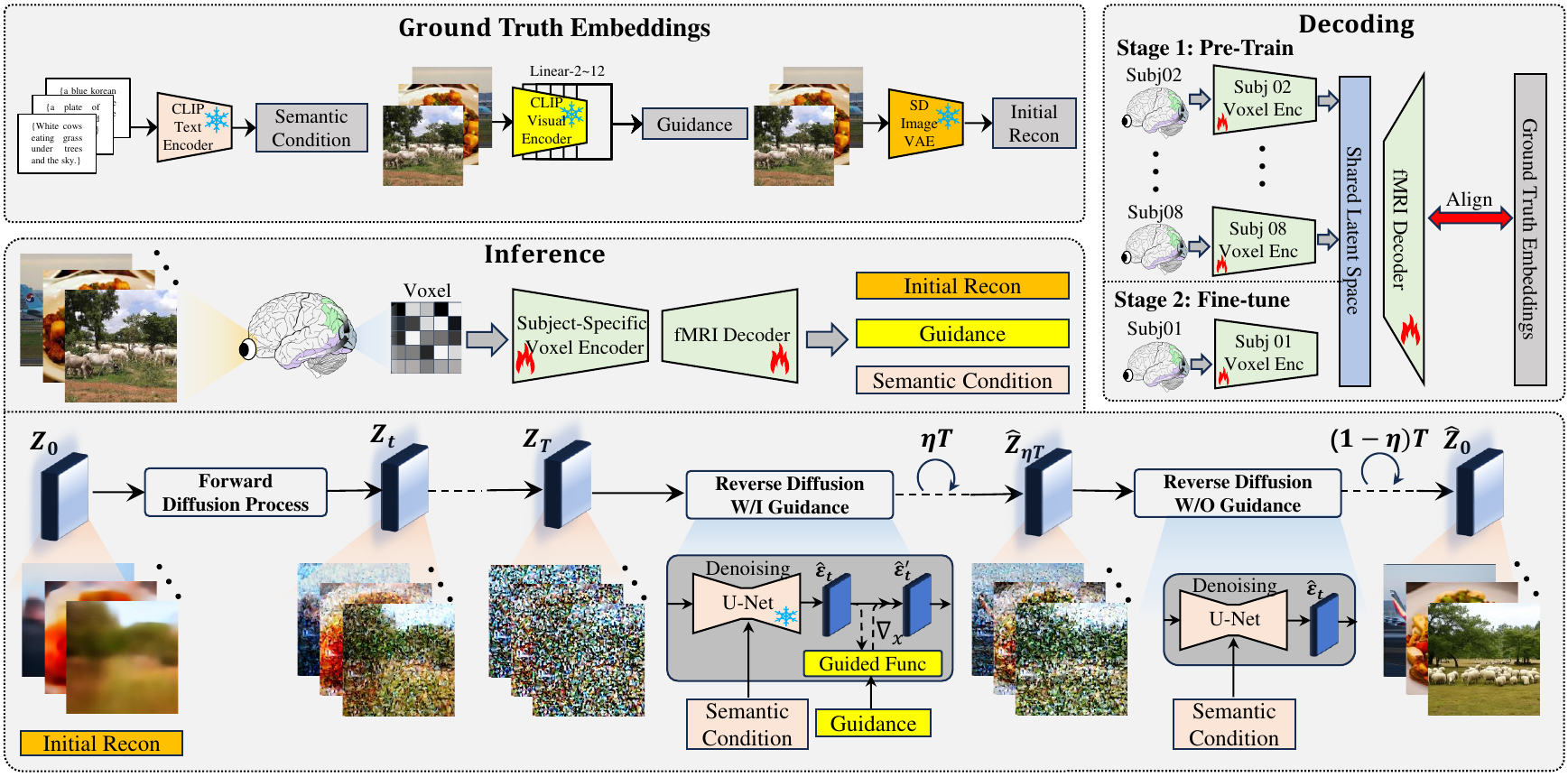}}
    \end{center}
   \caption{Overview of \systemname. During training, fMRI voxels are first mapped to a subject-shared space and then aligned with ground truth embeddings, namely CLIP text space, VAE's latent space and multilayers of CLIP image encoder. During inference, models run a forward and reverse diffusion process with or without guidance to generate reconstructed images.}
    \label{fig:framework}
\end{figure*}

\subsection{Overview}
Below, we elaborate on the details of the methods involved in our proposed \systemname. Figure \ref{fig:framework} presents the architecture of \systemname. Visual reconstruction seeks to map brain voxels $\mathcal{V}=\{v_i\}_{i=1}^N$ to visual images $\mathcal{X}=\{x_i\}_{i=1}^N$. Benefiting from LDM's well-mapping from text/image to image, the fMRI reconstruction task is simplified to align fMRI and LDM embeddings, that is $f_z(\cdot): \mathcal{V} \rightarrow \mathcal{Z}$ and $f_c(\cdot): \mathcal{V} \rightarrow \mathcal{C}$, where $\mathcal{Z}=\{z_i\}_{i=1}^N$ denotes the low-level initial latents and $\mathcal{C}=\{c_i\}_{i=1}^N$ represents the high-level semantic conditions. We introduce the details of mapping $\mathcal{V}$ to $\mathcal{Z}$ and $\mathcal{C}$ in Section \ref{fMRI Embeddings Decoding}. To enhance detail fidelity, we propose primary visual feature guidance where the latent diffusion is guided by a new mapping $f_g(\cdot): \mathcal{V} \rightarrow \mathcal{G}$, where $\mathcal{G}=\{g_i\}_{i=1}^N$ is designed to represent the primary visual feature. We describe the primary visual feature-guided diffusion reconstruction in Section \ref{Primary Visual Feature Guidance}.

\subsection{fMRI Embeddings Decoding}
\label{fMRI Embeddings Decoding}
Decoding fMRI embeddings involves finding mapping functions $f_z(\cdot): \mathcal{V} \rightarrow \mathcal{Z}$ and $f_c(\cdot): \mathcal{V} \rightarrow \mathcal{C}$. The $f_c(\cdot)$ and $f_z(\cdot)$ are composed of a subject-specific voxel encoder and an fMRI decoder. Voxels are first encoded into a subject-shared space by the subject-specific voxel encoder and then decoded by the fMRI decoder into target embeddings, namely low-level initial latent and high-level semantic condition.

\subsubsection{Subject-Specific Voxel Encoder}
FMRI measurements are 3-D voxels $\mathcal{V}\in \mathbb{R}^{h*w*c}$ where $h$, $w$, $c$ are the size of the registration space. We select regions of interest (ROI) on the cerebral cortex and flatten them to a vector $\mathcal{V}\in \mathbb{R}^{d_s}$. The number of voxels $d_s$ in ROIs is different for each subject $s$ due to the different sizes of the skull. To improve decoding performance and model generalization by efficiently leveraging fMRI-image pairs, we developed the subject-specific voxel encoder, a 1-layer fully connected neural network, to align different voxel sizes into a shared space $\mathbb{R}^d$ with 4096-dims at the front of the decoding model.  

\subsubsection{fMRI Decoder}
We develop two separate decoding pipelines for target embeddings: the high-level pipeline is aligned with the semantic conditions, while the low-level pipeline to the initial latents. 

\textbf{High-Level Decoding Pipeline: }
The semantic conditions $c_i$ are extracted from the text space of CLIP ViT/L-14 with a size of $\mathbb{R}^{77\times768}$, where 77 is the maximum token length and 768 denotes the feature dimension for each token. Text embeddings only represent high-level conceptual semantics without mixing in visual semantics~\cite{rossion2000functional} as image embeddings. Thus, we adopt CLIP text embeddings as high-level conditions because prominent high-level concepts are easier to decode from fMRI than image embeddings. Furthermore, the model for text embeddings with a smaller size of $\mathbb{R}^{77*768}$ performs better than image embeddings, which have a larger size of $\mathbb{R}^{256*768}$. We claim that text embeddings do not include visual semantics, which is compensated by the proposed guidance in Section \ref{Primary Visual Feature Guidance}. In this way, details are integrated into the low-level pipeline rather than the high-level pipeline. It aligns with the bottom-up and top-down processing mechanisms in neuroscientific theory and facilitates inverse retrieval and network dissection as streams are processed hierarchically, similar to the brain cortex.

The high-level decoding pipeline comprises a backbone module, a projector, and a diffusion prior module. The backbone module is a 4-layer multi-later perceptron (MLP)-Mixer, with the output layer being $77\times768$ in this paper for the clip text embedding. The projector is used for contrastive learning and is a fully connected network with 3 hidden layers. The diffusion prior, following the backbone module, is a conditional diffusion model similar to DALL-E2 used to reduce the modality discrepancy between fMRI and target embeddings.

In the first third of the training stage, contrastive learning uses CLIP loss with MixCo~\cite{kim2020mixco}.

\begin{equation}
    \label{eq:lmix}
    \begin{aligned}
    \mathcal{L}_{mix} &= -\frac{1}{N}
    \bigg[
    \sum_{i=0}^{N-1}
    \bigg(
        \lambda_i\log\frac{\exp(\frac{p_i^*\cdot c_i^*}{\tau})}{P_i}
        +(1-\lambda_i)\log\frac{\exp(\frac{p_i^*\cdot c_i^*}{\tau})}{P_i}
    \bigg) \\
    &+
    \bigg(
        \lambda_j\log\frac{\exp(\frac{p_j^*\cdot c_j^*}{\tau})}{C_j}
        +(1-\lambda_j)\log\frac{\exp(\frac{p_j^*\cdot c_j^*}{\tau})}{C_j}
    \bigg)
    \bigg] 
    \end{aligned}
\end{equation}
Here $\tau$ is a temperature hyperparameter, $p$ is the projector output following MLP and $'*'$ represents L2-normalization. Note that $P_i=\sum_{m=0}^{N-1}\exp(\frac{p_i^*\cdot c_m^*}{\tau})$ and $C_j=\sum_{m=0}^{N-1}\exp(\frac{p_m^*\cdot c_j^*}{\tau})$.

After one-third of the training, we replaced the CLIP loss with a soft CLIP loss where soft probability distributions have better supervision than hard labels. 
\begin{equation}
    \mathcal{L}_{soft} = -\frac{1}{N}\sum_{i,j=0}^{N-1}\biggl[\frac{\exp(\frac{c_i^*\cdot c_j^*}{\tau})}{\sum_{m=0}^{N-1}\exp(\frac{c_i^*\cdot c_m^*}{\tau})} 
     \cdot \log\frac{\exp(\frac{p_i^*\cdot c_j^*}{\tau})}{\sum_{m=0}^{N-1}\exp(\frac{p_i^*\cdot c_m^*}{\tau})}\biggr]
\end{equation}

The diffusion prior is inspired by DALL-E2~\cite{ramesh2022hierarchical} to bridge the modality gap between fMRI and CLIP text space. We execute a diffusion prior with 100 timesteps, and in this case, there are no learnable queries involved, as in~\cite{scotti2023reconstructing}. The total training loss is defined as follows:
\begin{equation}
    \mathcal{L}_c=\mathcal{L}_{mix|soft} + \alpha\cdot \mathcal{L}_{prior}
\end{equation}

\textbf{Low-Level Decoding Pipeline: }
The initial latents are extracted from the VAE's latent space, with a size of $\mathbb{R}^{4\times64\times64}$ which denotes low-level initial reconstructions. To map fMRI to VAE's latent space, $f_z(\cdot)$ uses an MLP backbone with 4 residual blocks that map fMRI to a space with a size of $\mathbb{R}^{16\times16\times64}$. Subsequently, an upsampler, designed with a similar architecture to the VQ-VAE decoder, is adopted to upsample backbone outputs to $\mathbb{R}^{64\times64\times4}$, which matches the size of the initial latent. Similar to $f_c(\cdot)$, the MLP backbone of $f_z(\cdot)$ is followed by a projector that maps the output to $\mathbb{R}^{512\times16\times16}$. The loss function of $f_z(\cdot)$ consists of soft CLIP loss and Mean Absolute Error (MAE) loss:
\begin{equation}
    \mathcal{L}_z=\mathcal{L}_{mae}+\beta\cdot \mathcal{L}_{soft}
\end{equation}
\begin{equation}
    \mathcal{L}_{mae}=\frac{1}{N}\sum_{i=0}^{N-1}\big|\hat{z}_i-z_i\big|
\end{equation}

Moreover, a first-order momentum alignment is proposed to align the statistics (mean and standard deviation) of the initial latent between the training and test sets. We emphasize that this contributes to aligning the distribution centers of the source and target domains. First, we normalize initial latent $\hat{z}$ on test set:
\begin{equation}
    \hat{z}'=\frac{\hat{z}-\mu_{\hat{z}}}{\sigma_{\hat{z}}}
\end{equation}
Next, $\hat{z}'$ are aligned to the mean and standard deviation of the training set:
\begin{equation}
    \hat{z}=\sigma_{tr}\cdot \hat{z}'+\mu_{tr}
\end{equation}
where $\mu_{tr}$ and $\sigma_{tr}$ are the mean and standard deviation of the training set.

\subsubsection{Model Training: }
We trained subject-specific models for each subject, with the training process comprising two stages. First, fMRI-image pairs from the remaining subjects (leave-a-target-subject-out) were utilized for model pre-training to learn the implicit mapping between shared space and the target embedding. Second, we fine-tuned the model for the target subject. Two fine-tuning parameters were employed: 1) the target subject-specific voxel encoder and 2) the target-independent mapping between shared space and target embeddings. 

\subsection{Primary Visual Feature Guidance}
\label{Primary Visual Feature Guidance}
The LDM generates the reconstructed images using the decoded embeddings $\hat{z}$ and $\hat{c}$. The output images exhibit semantic similarities to the ground truth, but the detail fidelity is limited. We attribute the suboptimal results to the lack of detailed information and the diverse outputs of the generative model. Here, we propose primary visual feature guidance to address this challenge.

\textbf{Universal Guided Reconstruction: }
Score-based diffusion model derived in ~\cite{song2021scorebased} defines a stochastic differential equation for the reverse diffusion process:
\begin{equation}
    dz = \big[f(z,t)-g^2(t)s(z_t)\big]dt+g(t)d\bar{w}\\
\end{equation}
where $s(z_t) = \nabla_{z_t}\log{p(z_t)}$ is a score function that defines the generative tendency. Considering a conditional probability $p(z_t|y)$, the score function $s(z_t,y)$ can be derived by Bayes' rule as follows:
\begin{equation}
    \begin{aligned}
        s(z_t,y)&=\nabla_{z_t}\log{p(z_t|y)}\\
        &=\nabla_{z_t}\log{p(z_t)}+\nabla_{z_t}\log{p(y|z_t)}
    \end{aligned}
\end{equation}
where $\nabla_{z_t}\log{p(y|z_t)}$ defines a classifier guidance ~\cite{dhariwal2021diffusion}. We generalize the classifier $\log{p(y|z_t)}$ to a universal guidance function:
\begin{equation}
    \begin{aligned}
        s(z_t,g)&=\nabla_{z_t}\log{p(z_t)}-\nabla_{z_t}\mathcal{L}_{g}(z_t,g)
    \end{aligned}
\end{equation}

However, the $z_t$ in $\mathcal{L}_{g}(z_t,g)$ represents a noisy image associated with the $t$-th diffusion step, rather than a clear image. Therefore, the pre-trained visual model cannot be employed as a plug-and-play guidance function; otherwise, the generated images will be distorted by the high noise input. To implement plug-and-play guidance functions, we estimate the clear image from the noisy image using Tweedie's method~\cite{efron2011tweedie, kim2021noise2score} and obtain the final image using a weighted formula with Euler's formula approximation.
\begin{equation}
    \hat{z}_0:= \mathbb{E}[z_0|z_t] =\frac{1}{\sqrt{\bar{\alpha}_t}}(z_t+(1-\bar{\alpha}_t)s(z_t))
\end{equation}
\begin{equation}
    \hat{z}_t=\sqrt{1-\bar{\alpha}_t}\hat{z}_0+(1-\sqrt{1-\bar{\alpha}_t})z_t
\end{equation}

\textbf{Primary Visual Feature Decoding: }
A plethora of guidance functions have been developed in the domain of computer vision (CV), encompassing such diverse applications as skeleton, object detection, and semantic segmentation. However, the process of decoding these aforementioned guidance functions from fMRI data has proven to be challenging, resulting in suboptimal accuracy and the necessity for complex decoders. Inspired by the findings reported in~\cite{yang2023brain, Wang2023NaturalLS}, we have observed that deep models exhibit a hierarchical structure that bears a striking resemblance to that of the human brain. Recent research~\cite{yang2023brain} compared the contributions of multiple feature layers from various deep models when predicting brain fMRI responses and found that among the various deep models (such as CLIP, ImgNet, SAM, etc.), the CLIP image encoder is the most similar to the hierarchical structure of the human brain. Therefore, we extract multilayers (layer-2,4,6,8,10,12) of CLIP image encoder with size of $\mathbb{R}^{50*768}$ as guidance:
\begin{equation}
    g_i^l = CLIP_{image}(x_i, l),\quad l=2,4,6,8,10,12
\end{equation}
where $l$ is the $l$-th layer of CLIP image encoder. 

To decode $\mathcal{G}^l=\{g_i^l\}_{i=1}^N$ from fMRI, $f_g^l(\cdot)$ adopts the same architecture as high-level pipeline and the output size of the MLP backbone is set to $\mathbb{R}^{50\times768}$ instead of $\mathbb{R}^{77\times768}$. Model training follows the same procedure as the high-level pipeline.

\textbf{Primary Visual Feature-Guided Reconstruction: }
We aim to reconstruct the visual stimuli rather than generate a new image, a significant departure from the original intention of guided diffusion in the computer vision community. To this end, we propose a novel guidance strategy. Firstly, a hyperparameter, the guidance scale $\kappa$, is set to adjust the weight of guidance. A large scale of guidance $\kappa=300,000$ is implemented as a strong constraint to highlight the depiction of details by primary visual features.
\begin{equation}
    \mathcal{L}_{g}(z_t,g)=\kappa\cdot \sum_{l\in L}^{}{||f_g^l(\mathcal{D}(\hat{z}_t))-\hat{g}^l||_2^2}
\end{equation}
where $\mathcal{D(\cdot)}$ is the VQ-VAE's decoder.

However, the primary visual feature guidance decoded from fMRI inevitably introduces a significant number of decoding errors, which are simultaneously highlighted by large guidance scales.
In practice, we observed that reconstructed images exhibit artifacts and blur (see Fig.\ref{fig:guidance_param}). Our solution is based on the observation that the guidance focuses on the early steps of the reverse diffusion process. We introduce a hyperparameter (guidance strength $\eta$) to control the guided steps. \systemname\ adopts a small $\eta=0.2$ since the latent $\hat{z}_t$ is corrected within few guided steps. The total guided diffusion algorithm for reconstruction tasks is outlined in Algorithm \ref{alg:guidance}.

 \begin{algorithm}[!t]
    \caption{primary visual feature guidance algorithm}
    \label{alg:guidance}
    \SetAlgoLined
    {\bfseries Input:} $\hat{z}_i$, $\hat{c}_i$, $\hat{g}_i$, $\kappa$, $\eta$\;
    \textbf{Initialize: }$z_0\leftarrow \hat{z}_i$, $z_T\leftarrow \sqrt{\bar{a}_T}z_{0}+\sqrt{1-\bar{a}_T}\epsilon$ \\
    \setstretch{1.3}
    \For {$t=T\rightarrow1$}{
        $\hat{z}_0=\frac{1}{\sqrt{\bar{\alpha}_t}}(z_t+(1-\bar{\alpha}_t)\nabla_{z_t}\log{p(z_t)})$\\
        \eIf{$t<\eta T$}{
                $\hat{z}_t=\sqrt{1-\bar{\alpha}_t}\hat{z}_0+(1-\sqrt{1-\bar{\alpha}_t})z_t$\\
                $\mathcal{L}_{g}(z_t,g)=\kappa\cdot \sum_{l\in L}^{}{||f_g^l(\mathcal{D}(\hat{z}_t))-\hat{g}^l||_2^2}$\\
                $\hat{\epsilon}=\epsilon_\theta(z_t, \hat{c})+\sqrt{1-\bar{\alpha}_t}\nabla_{z_t}\mathcal{L}_{g}(\hat{z}_t,\hat{g})$\\
                $z_{t-1}=\sqrt{\bar{\alpha}_{t-1}}\bigl(\frac{z_t-\sqrt{1-\bar{\alpha}_t}\hat{\epsilon}}{\sqrt{\bar{\alpha}_t}}\bigr)+\sqrt{1-\bar{\alpha}_{t-1}}\hat{\epsilon}$\\
        }{
            $z_{t-1}=\sqrt{\bar{\alpha}_{t-1}}\hat{z}_0+\sqrt{1-\bar{\alpha}_{t-1}}\epsilon_\theta(z_t, \hat{c})$\
        }
    }
 \end{algorithm}

\section{Experiments}
\label{Experiments}
\subsection{Dataset and Setup}
\label{Dataset and Setup}
\textbf{Dataset:} We evaluate \systemname\ using the Natural Scenes Dataset (NSD)~\cite{allen2022massive}, which comprises a total of 73,000 images derived from the Common Objects in Context (COCO) dataset~\cite{lin2014microsoft}. These images were viewed by 8 subjects, with each subject being assigned 10,000 images, including 9,000 unique images and 1,000 shared images. Each subject was presented with 10,000 images, repeated 3 times in 30-40 sessions during 7T fMRI scanning. We evaluated four subjects (subj01, subj02, subj05, and subj07) who completed all imaging sessions. Scanning was conducted at a resolution of 1pt8mm, and voxels in the \texttt{nsdgeneral} ROI provided by the dataset were selected. We followed the standard data split used in previous studies, with 1,000 shared images designated for the test set and 9,000 unique images for model training. We averaged across 3 repeated scans in the test set (1,000 trials) but not in the training set (27,000 trials).

\textbf{Metrics:} Following previous work~\cite{kneeland2023brain, scotti2024mindeye2}, we employ a set of 8 image metrics for quantitative evaluation, including pixel-level correlation (PixCorr) and Structural Similarity Index (SSIM)~\cite{wang2004image}. AlexNet(2) and AlexNet(5) represent the two-way identification of the 2nd and 5th feature layers of AlexNet~\cite{krizhevsky2012imagenet}, respectively. CLIP and IncepV3 denote the two-way identification of the CLIP ViT/L-14 model~\cite{radford2021learning} and the last pooling layer of Inception V3~\cite{szegedy2016rethinking}, respectively. Two-way identification is defined as the percentage that measures the Pearson correlation between reconstructions and the ground truth, comparing whether the embeddings of the reconstruction are more similar to those of the ground truth than to randomly selected ones. We followed the implementation of Takagi \emph{et al.}~\cite{takagi2023high}. EffNet-B~\cite{tan2019efficientnet} and SwAV~\cite{caron2020unsupervised} are distance metrics based on EfficientNet-B13 and SwAV-ResNet50, respectively.

\begin{table*}[!t]
    \centering
    \caption{Comparison of metrics with the state-of-the-art and other guided methods. \textbf{Best} and \underline{Second} are highlighted.}
    \label{tab:compare_sota}
    \resizebox{1.\linewidth}{!}{
    \begin{tabular} {ll|cccc|cccc|cc}
        \toprule  
        \textbf{Method} & & \multicolumn{4}{c|}{\textbf{Low-Level}} & \multicolumn{4}{c|}{\textbf{High-Level}} & \multicolumn{2}{c}{\textbf{Retrieval}} \\
        & & SSIM$\uparrow$ & PixCorr$\uparrow$ & AlexNet(2)$\uparrow$ & AlexNet(5)$\uparrow$ & IncepV3$\uparrow$ & CLIP$\uparrow$ & EffNet-B$\downarrow$ & SwAV$\downarrow$ & Image$\uparrow$ & Brain$\uparrow$  \\
        \midrule
        \multirow{1}{*}{{GANs}}& {MindReader (NeurIPS'22)} & -- & -- & -- & -- & {78.2\%} & -- & -- & -- & {11.0\%} & {49.0\%} \\
        & {Gu et al. (MIDL'23)} & {0.325} & {0.150} & -- & -- & -- & -- & {0.862} & {0.465} & -- & -- \\
        \midrule
        \multirow{1}{*}{{\makecell{Stable-\\Diffusion}}} & Takagi \emph{et al.} (CVPR'23) & -- & -- & 83.0\% & 83.0\% & 76.0\% & 77.0\% & -- & -- & -- & -- \\
        & BrainDiffuser (Sci-Rep'23) & \textbf{0.356} & 0.254 & 94.2\% & 96.2\% & 87.2\% & 91.5\% & 0.775 & 0.423 & 21.1\% & 30.3\% \\
        & MindEye (NeurIPS'23) & 0.323 & \textbf{0.309} & \underline{94.7\%} & \underline{97.8\%} & 93.8\% & \underline{94.1\%} & \textbf{0.645} & \textbf{0.367} & \underline{93.6\%} & \underline{90.1\%} \\
        \midrule
        \multirow{1}{*}{{Optimized}} & MindDiffuser (ACM MM'23) & \underline{0.354} & 0.278 & -- & -- & -- & 76.5\% & -- & -- & -- & -- \\
        & MindEye+BOI (ArXiv'23) & 0.329 & 0.259 & 93.9\% & 97.7\% & \underline{93.9\%} & 93.9\% & \textbf{0.645} & \textbf{0.367} & -- & -- \\
        & DREAM (WACV'24) & 0.338 & 0.288 & 93.9\% & 96.7\% & 93.7\% & \underline{94.1\%} & \textbf{0.645} & 0.418 & -- & -- \\
        \midrule
        \textbf{Ours} & {\textbf{\systemname}} & 0.318 & \underline{0.299} & \textbf{95.3\%} & \textbf{98.3\%} & \textbf{95.4\%} & \textbf{95.4\%} & \underline{0.663} & \underline{0.401} & \textbf{98.81\%} & \textbf{97.54\%} \\
        \bottomrule
    \end{tabular}
    }
\end{table*}

\begin{figure}[!t]
    \begin{center}
        \centerline{\includegraphics[width=.9\linewidth]{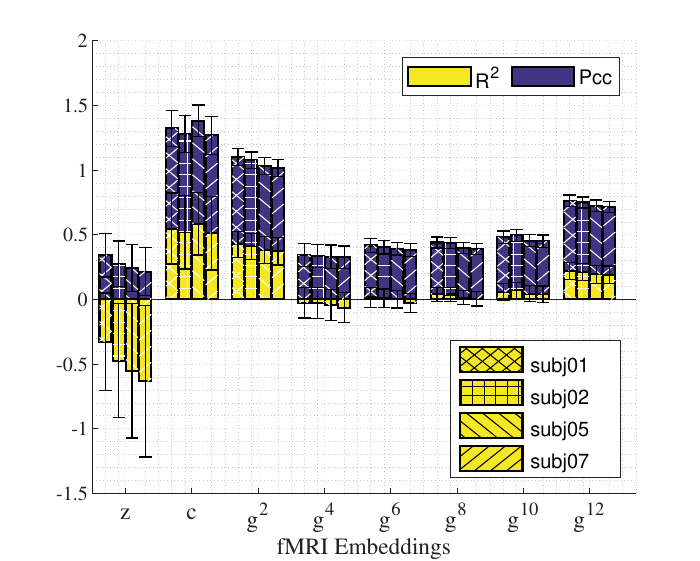}}
    \end{center}
    \caption{Decoding performance of fMRI embeddings.}
    \label{fig:fmri_decoding}
\end{figure}

\begin{figure*}[!t]
    \begin{center}
        \centerline{\includegraphics[width=1\linewidth]{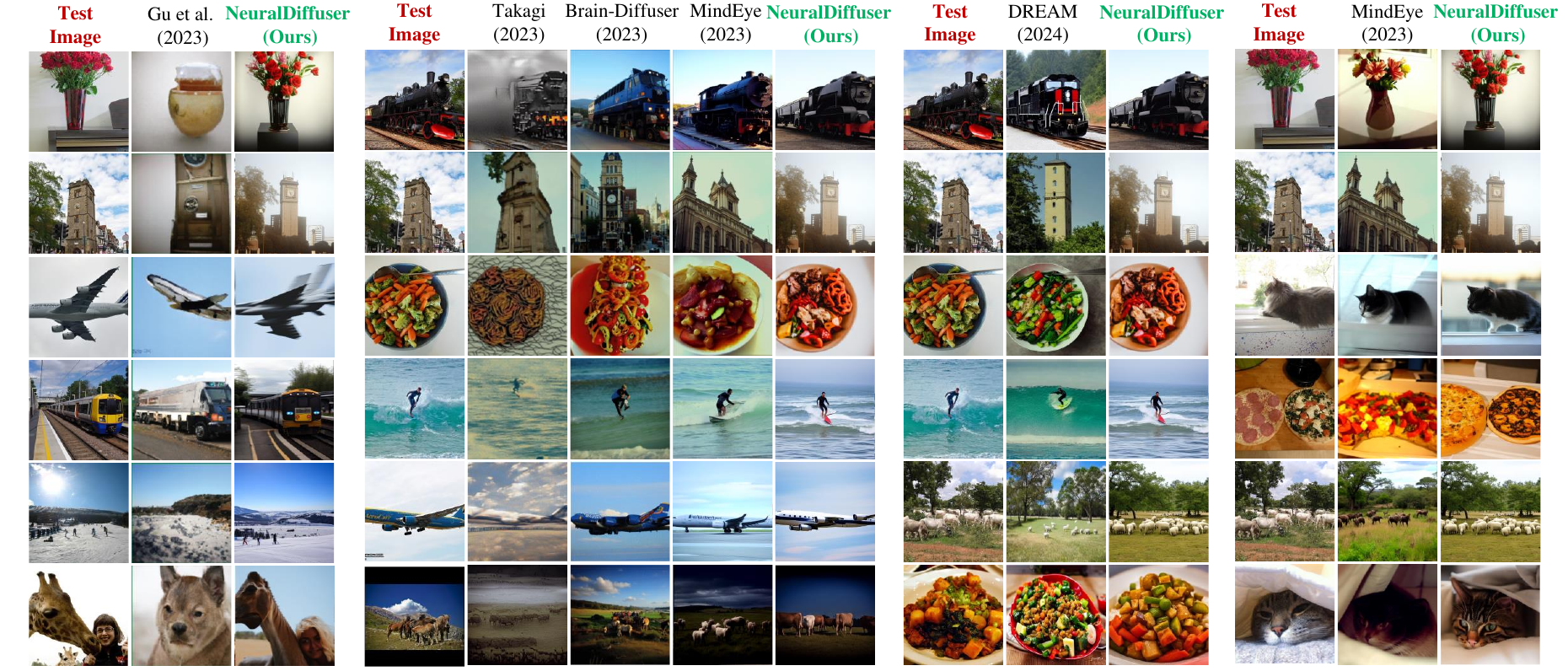}}
    \end{center}
   \caption{
    {Reconstructed images comparison for \systemname\ against the state-of-the-art}}
    \label{fig:compare_sota}
\end{figure*}

\textbf{Implementation:} Pre-training and fine-tuning are implemented on 6 \texttt{NVIDIA GeForce RTX 3090} GPUs for 150 epochs. The hyperparameter $\alpha$ is set to 30 for training the $f_c$ and $f_g$ models. The batch size is set to 21, and the optimizer is configured to AdamW with a cycle learning rate schedule, where the maximum learning rate is set to $3 \times 10^{-4}$. We utilize \texttt{stable-diffusion-v1-4} to generate images with a size of $512 \times 512$. We use 50 steps of \texttt{DDIMSampler} with a reverse diffusion strength of 0.75 (37 steps). The classifier-free scale is set to 7.5 and the self-attention guidance (SAG) scale is set to 0.75 by default. We set the primary visual feature guidance hyperparameters $\kappa=300,000$ and $\eta=0.2$ unless otherwise specified. As guidance involves gradient propagation, we recommend a \texttt{GPU VRAM} of at least 24 GB, such as the \texttt{NVIDIA GeForce RTX 3090}. It should be noted that guidance operates during the initial 20\% ($\eta=0.2$) of the reverse diffusion steps and therefore does not lead ti a significant increase in the time cost.

\begin{figure*}[!t]
    \begin{center}
        \centerline{\includegraphics[width=1\linewidth]{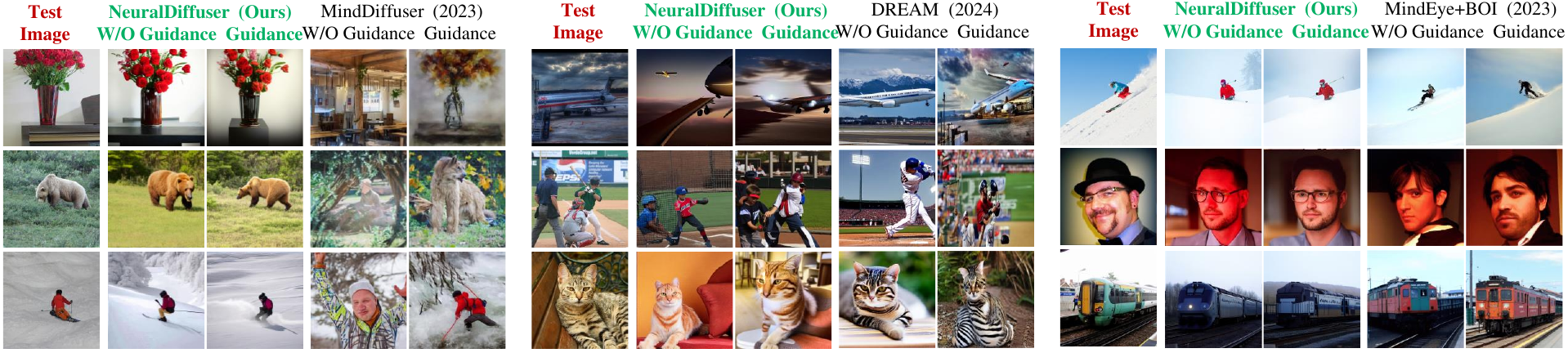}}
    \end{center}
   \caption{Reconstructed images comparison for \systemname\ against other guided methods}
    \label{fig:compare_guided}
\end{figure*}

\subsection{fMRI Embeddings Decoding}
\label{Feature Decoding Experiments}
Decoding fMRI embeddings $\hat{z}$, $\hat{c}$, and $\hat{g}$ is a crucial step in visual reconstruction. We evaluate the decoding performance using the Pearson correlation coefficient and $R^2$ values. Fig.~\ref{fig:fmri_decoding} illustrates the decoding performance of fMRI embeddings, comprised of hierarchical features aligned with stable diffusion models and guidance functions. Specifically, the following components are incorporated: 1) VQ-VAE embeddings ($z$): low-level details; 2) CLIP text embeddings ($c$): high-level semantics; and 3) guidance features ($g^{2-12}$): the multilayers of the CLIP image encoder. Each bar represents the average and standard deviation of the Pearson correlation coefficient (in blue) and $R^2$ values (in yellow) derived from the 1,000 test samples for each subject and embedding. The results indicate that the high-level decoding pipeline demonstrates better performance for the CLIP text embeddings (semantic conditions $c$), while the low-level pipeline struggles with the VQ-VAE embeddings (initial latent $z$). This finding suggests that high-level semantics can be effectively decoded from fMRI data, whereas low-level details pose significant challenges. Consequently, the reconstructed images based on stable diffusion are semantically coherent but often lack detail fidelity.

Note that the suboptimal performance of the decoding of $z$ is not unique to \systemname, but rather a common issue faced in previous studies, often resulting in unfaithful details. In fact, the latent space of VQVAE is a compressed representation of images in a low-dimensional space. Therefore, decoding $z$ is equivalent to pixel-level visual reconstruction, which often results in blurry images, as seen in previous work. To address this challenge, \systemname\ employs primary visual feature guidance to provide detailed cues. The decoding performance of guidance $g$, which uses multilayer features from CLIP, demonstrating a significant enhancement in decoding accuracy when compared to the latent space of the VAE.

\subsection{Retrieval}
Retrieval involves selecting the most similar samples from a given set. The retrieval method is derived from MindEye~\cite{scotti2023reconstructing} and includes both image retrieval and brain retrieval. 
1). Image Retrieval: Given an fMRI, we decode it into an fMRI embedding and select the image from 300 random images in the test set based on the highest cosine similarity between the fMRI embedding and the 300 image embeddings (also known as top-1 retrieval performance; 300 candidates, chance = $\frac{1}{300}$). 
2). Brain Retrieval: Given an image, we select the fMRI embedding that exhibits the highest cosine similarity from the 300 random fMRIs in the test set. If the fMRI and the image match, the retrieval is considered correct; otherwise, it is deemed incorrect. 
We implemented a retrieval task using the projector output of the fMRI decoder and calculated the average retrieval accuracy of 1,000 fMRI-image pairs as our evaluation metric.

MindEye~\cite{scotti2023reconstructing} retrieves based on a high-level pipeline (representing fMRI image embeddings). However, the retrieval rate based on the high-level pipeline only achieved 49.75\% for image retrieval and 48.23\% for brain retrieval. In \systemname, the high-level pipeline aims to align with the CLIP text space, which is less dense than the CLIP image space. For instance, the phrase "a giraffe on the grass" can be interpreted in a variety of ways, making it challenging to retrieve the intended image based solely on the fMRI text embedding. We hypothesize that effective retrieval requires integration with visual semantics (that is, primary visual features). Guided by this hypothesis, we have developed a retrieval method based on primary visual features. Specifically, the guidance features (multi-layer features of the CLIP image encoder) are used to retrieve images and brain fMRIs separately and are then merged using a voting strategy. The retrieval performance of \systemname\ compared to that of recent studies in the bottom of Tab.~\ref{tab:compare_sota}. 
The focus on decoding detailed cues from fMRI as guidance provides advantages for bidirectional retrieval conditioned on primary visual features, thus achieving a remarkable improvement in retrieval performance.

\subsection{Image Reconstruction}
\textbf{Comparison with State-of-the-Art:} We report a comparison of \systemname\ against several recent studies, namely MindReader~\cite{lin2022mind}, Gu \emph{et al.}~\cite{gu2023decoding}, Takagi \emph{et al.}~\cite{takagi2023high} (LDM-based reconstruction baseline), Brain-Diffuser~\cite{ozcelik2023natural}, DREAM~\cite{xia2024dream}, and MindEye~\cite{scotti2023reconstructing}. Fig.~\ref{fig:compare_sota} presents a comparison of several reconstructed images generated by \systemname\ against the state-of-the-art. Images are taken from the original papers if reported or from our reproduction. A comparative analysis reveals that \systemname\ outperforms that of alternative methods, particularly in terms of elements such as background, structure, color, and texture.

In the qualitative evaluation, we followed the reproducibility results reported in MindEye~\cite{scotti2023reconstructing}. For each test image, five images were generated and the comparison results were reported, averaged across five replicates and four subjects in Tab.~\ref{tab:compare_sota} (top). A comparative analysis of the metrics reveals the superiority of \systemname\ over the baseline approaches. We find that the two-way identification metrics (AlexNet(2), AlexNet(5), CLIP, InceptionV3) are significantly improved, while the similarity (SSIM and Pixcorr) and distance (EffNet-B and SwAV) metrics remain suboptimal. This suboptimality may be attributed to the reporting of the average of five replicates rather than the best reconstruction through second-order selection. The best values obtained from five repetitions indicate that SSIM reaches 0.34, Pixcorr up to 0.343, EffNet-B achieves a score of 0.58, and SwAV is 0.346, which are considerably superior to the baseline.

\begin{figure*}[!t]
    \begin{center}
        \centerline{\includegraphics[width=1\linewidth]{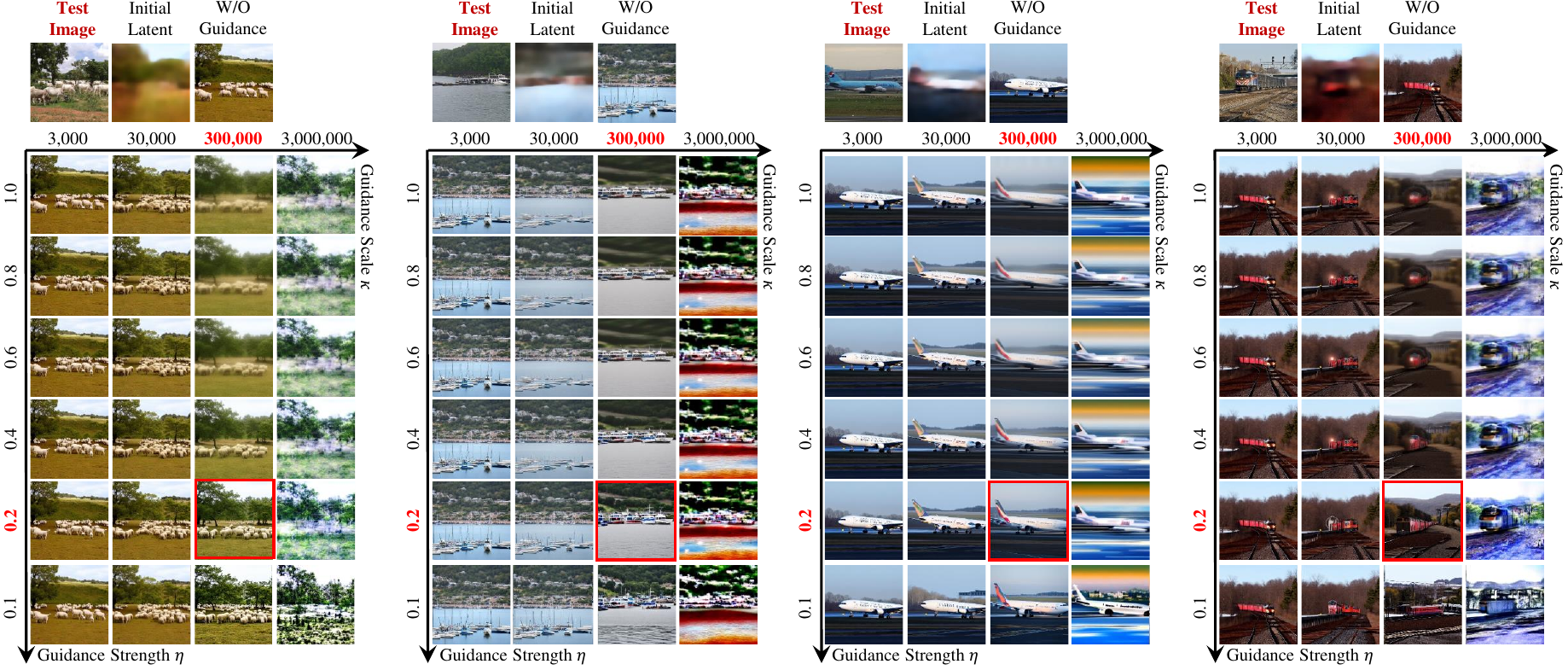}}
    \end{center}
   \caption{Reconstructed images comparison of different guidance hyperparameters}
    \label{fig:guidance_param}
\end{figure*}

\begin{table*}[!t]
    \centering
    \caption{Evaluated metrics of guidance hyperparameters (scale \& strength). \textbf{Best} and \underline{Second} are highlighted.}
    \label{tab:guidance_param}
    \resizebox{1.\linewidth}{!}{
    \begin{tabular} {c|c|ccc|cccccc|c}
        \toprule  
        \multirow{2}{*}{Parameter} & guidance scale $\kappa$ & 0 & 3,000 & 30,000 & \multicolumn{6}{c|}{\textbf{300,000}} & 3,000,000 \\ 
        & guidance strength $\eta$ & -- & 1.0 & 1.0 & 1.0 & 0.8 & 0.6 & 0.4 & \textbf{0.2} & 0.1 & 1.0 \\ 
        \midrule
        \multirow{4}{*}{Low-Level} & SSIM$\uparrow$ & 0.304 & 0.306 & 0.325 & \textbf{0.371} & \textbf{0.371} & \underline{0.368} & 0.352 & 0.318 & {0.290} & 0.296 \\
        & PixCorr$\uparrow$ & 0.294 & 0.295 & 0.301 & \textbf{0.319} & \textbf{0.319} & \underline{0.317} & 0.311 & 0.299 & {0.286} & 0.217 \\
        & Alexnet(2)$\uparrow$ & 92.39\% & 92.63\% & 93.99\% & 95.1\% & 95.13\% & 95.16\% & \underline{95.24\%} & \textbf{95.33\%} & {94.61\%} & 93.92\% \\
        & Alexnet(5)$\uparrow$ & 96.62\% & 96.72\% & 97.39\% & 97.71\% & 97.62\% & 97.67\% & 97.96\% & \textbf{98.28\%} & {\underline{98.06\%}} & 97.17\% \\
        \midrule
        \multirow{4}{*}{High-Level} & CLIP$\uparrow$ & 95.01\% & 94.99\% & 95.31\% & \textbf{96.09\%} & \underline{95.93\%} & 95.83\% & 95.74\% & 95.41\% & {95.08\%} & 86.28\% \\ 
        & Incep-V3$\uparrow$ & 94.71\% & 94.6\% & 94.82\% & 95.08\% & 94.87\% & 94.93\% & \underline{95.21\%} & \textbf{95.37\%} & {95.27\%} & 89.78\% \\
        & EffNet-B$\downarrow$ & 0.668 & 0.667 & \textbf{0.661} & 0.735 & 0.742 & 0.735 & 0.698 & \underline{0.663} & {0.669} & 0.877 \\
        & SwAV$\downarrow$ & 0.413 & 0.411 & \underline{0.403} & 0.485 & 0.501 & 0.495 & 0.449 & \textbf{0.401} & {0.402} & 0.601 \\
        \midrule
        \multirow{6}{*}{Brain Corr.}  & V1$\uparrow$ & 0.286 & 0.290 & 0.317 & 0.340 & 0.328 & 0.330 & 0.350 & \underline{0.357} & {0.321} & \textbf{0.386} \\
        & V2$\uparrow$  & 0.257 & 0.260 & 0.287 & 0.301 & 0.290 & 0.294 & 0.317 & \underline{0.326} & {0.297} & \textbf{0.358} \\
        & V3$\uparrow$  & 0.248 & 0.252 & 0.275 & 0.279 & 0.271 & 0.275 & 0.299 & \underline{0.310} & {0.285} & \textbf{0.330} \\
        & V4$\uparrow$  & 0.249 & 0.251 & 0.270 & 0.255 & 0.249 & 0.254 & 0.279 & \textbf{0.296} & {0.281} & \underline{0.289} \\
        & Higher Vis.$\uparrow$  & 0.322 & 0.324 & 0.335 & 0.296 & 0.291 & 0.299 & \underline{0.328} & \textbf{0.346} & {0.343} & 0.292 \\
        & Whole Vis.$\uparrow$  & 0.316 & 0.318 & 0.333 & 0.303 & 0.294 & 0.301 & \underline{0.333} & \textbf{0.353} & {0.292} & 0.324 \\
        \midrule
        \multirow{2}{*}{Quality} & IS$\uparrow$ & \textbf{13.62} & \underline{13.61} & 13.45 & 10.10 & 9.61 & 9.78 & 11.1 & 12.55 & {13.19} & 5.70 \\
        & FID$\downarrow$ & \underline{47.61} & \textbf{47.54} & 48.40 & 73.20 & 73.87 & 71.14 & 58.11 & 50.42 & {47.89} & 138.54 \\
        \bottomrule
    \end{tabular}
    }
\end{table*}

Tab.~\ref{tab:compare_sota} (bottom) presents a comparison of \systemname\ against several works dedicated to improving detail fidelity, including MindDiffuser~\cite{lu2023minddiffuser}, DREAM~\cite{xia2024dream}, and MindEye+BOI~\cite{kneeland2023brain}. Our proposed \systemname\ exhibits comparable performance in low-level details in both qualitative observations and quantitative metrics. In terms of time cost, although computing gradients consumes time during inference, \systemname\ guides the diffusion process in a limited number of steps (20\% in the paper), which is comparable to DREAM and superior to MindDiffuser and MindEye+BOI, which requires iterative diffusion. Furthermore, \systemname\ proposes a universal guidance method that is not limited to CLIP features. It is noteworthy that alternative guidance functions, such as those implemented in MindDiffuser, DREAM, and MindEye+BOI, are also deemed acceptable.

\subsection{Ablation Study}
\label{Ablation Study}
In this section, we present the ablation study of the proposed primary visual feature guidance. Additionally, we discuss the hyperparameters of the guidance scale $\kappa$ and guidance strength $\eta$ used in the guidance algorithm.

\subsubsection{Guidance hyperparameters}
Our proposed guidance strategy sets two hyperparameters: the guidance scale $\kappa$, which exerts an influence on the weight of details, and the guidance strength $\eta$, which governs the number of guidance steps. In this study, we change the two hyperparameters $\kappa$ and $\eta$ and present the reconstructed results. It is imperative to note that the guidance scale, $k=0$, corresponds to a reconstruction w/o guidance.
\begin{figure*}[!t]
    \begin{center}
        \centerline{\includegraphics[width=1\linewidth]{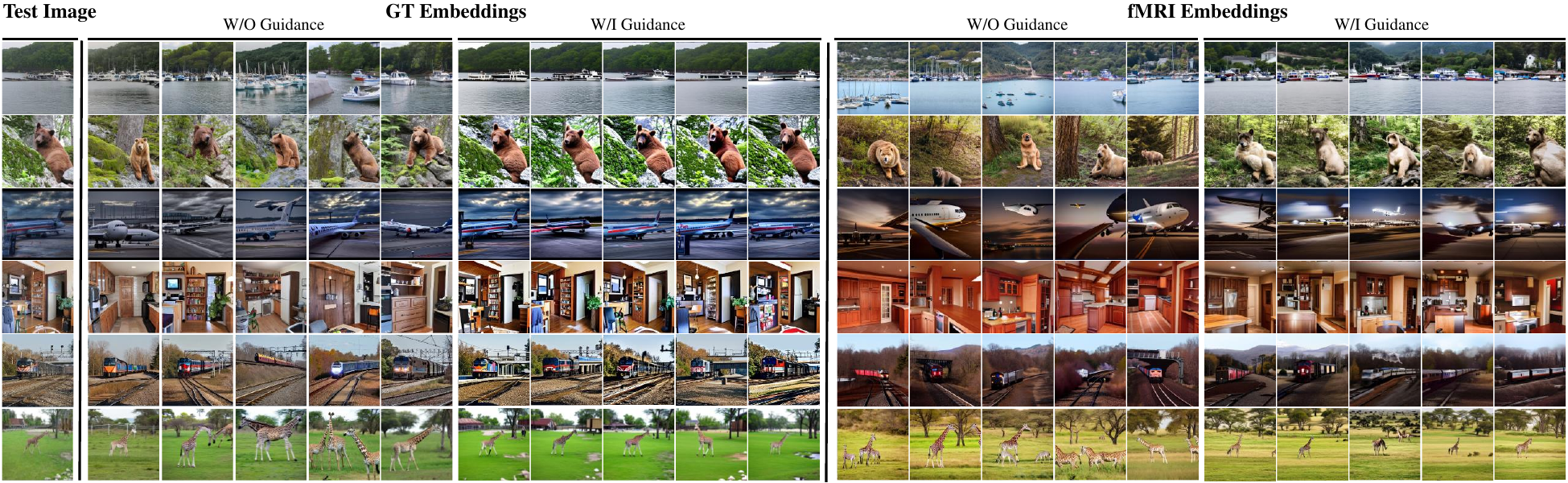}}
    \end{center}
   \caption{Repeatedly reconstruct 5 images by ground truth (GT) embeddings and fMRI embeddings}
    \label{fig:onetomany}
\end{figure*}

We present reconstruction examples of four test images with different hyperparameters in Fig.~\ref{fig:guidance_param} and compare the evaluated metrics of different hyperparameters in Tab.~\ref{tab:guidance_param}. The brain correlation scores are used to evaluate how well the reconstructed images can predict brain voxel activation using fMRI encoding models. Specifically, we utilized GNet~\cite{st2022brain}, a pre-trained fMRI encoding model, to predict the fMRI responses induced by the reconstructed images. We then analyzed the Pearson correlation coefficient between the brain responses simulated from \systemname's reconstructed images for the 1,000 test samples and the ground truth fMRI measurements. The terms V1, V2, V3, and V4 refer to the brain correlation scores associated with relevant cortical regions of interest (ROIs). Additionally, "Higher Vis" denotes the scores related to higher visual areas, while "Whole Vis" refers to the correlation scores for the entire visual cortex. To assess the quality of the reconstructions, two widely employed metrics are utilized: Inception Score (IS)~\cite{salimans2016improved} and Fréchet Inception Distance (FID)~\cite{Seitzer2020FID}.

Results without guidance (w/o: $\kappa = 0$) lead to the conclusion that uncontrolled diffusion models tend to fail to the details of reconstructed images because the blurred initial latent cannot provide adequate detailed cues. 
Our findings indicate that while increasing the guidance scale improves low-level detail accuracy, it concomitantly leads to a diminution in overall image quality, as evidenced by IS and FID metrics. This is attributable to the fact that guidance features are not predefined as in computer vision but are decoded from fMRI data, where the detailed information is compromised by decoding errors. 
The investigation unveils the guided diffusion process, elucidating that the guidance is designed to correct detailed trends originating from the blurred initial latent and finishing within a few steps. This finding led to the establishment of a strength hyperparameter to adjust the guidance steps. From a vertical perspective, lower strength enhances image clarity and quality without compromising detail fidelity. 
The total diffusion process comprises two distinct stages: 1) the correction of details in the first $\eta$ steps and 2) the enhancement of quality in the subsequent $1 - \eta$ steps. It is imperative to acknowledge the trade-off inherent in this process: an augmented guidance scale prioritizes detail fidelity, while a diminished guidance scale enhances image quality. The best performance is attained when setting $\kappa = 300,000$ and $\eta = 0.2$.

The findings from the ablation of the proposed guidance suggest that primary visual features play a pivotal role in visual reconstruction when employing stable diffusion models. 
Irrespective of whether model metrics or brain correlation scores are utilized, the outcomes substantiate the hypothesis that primary visual features prioritize enhancing the low-level details of reconstruction without compromising or even improving the high-level semantic coherence. 
The improvement of low-level metrics can be anticipated due to the fact that our proposed primary visual feature guidance provides detailed cues and controls the reverse diffusion process. 
The high-level improvements indicate that \systemname\ has the potential to extract high-level semantic knowledge from low-level details, reflecting a bottom-up process in neuroscience theory.

\begin{figure*}[!t]
    \begin{center}
        \centerline{\includegraphics[width=1\linewidth]{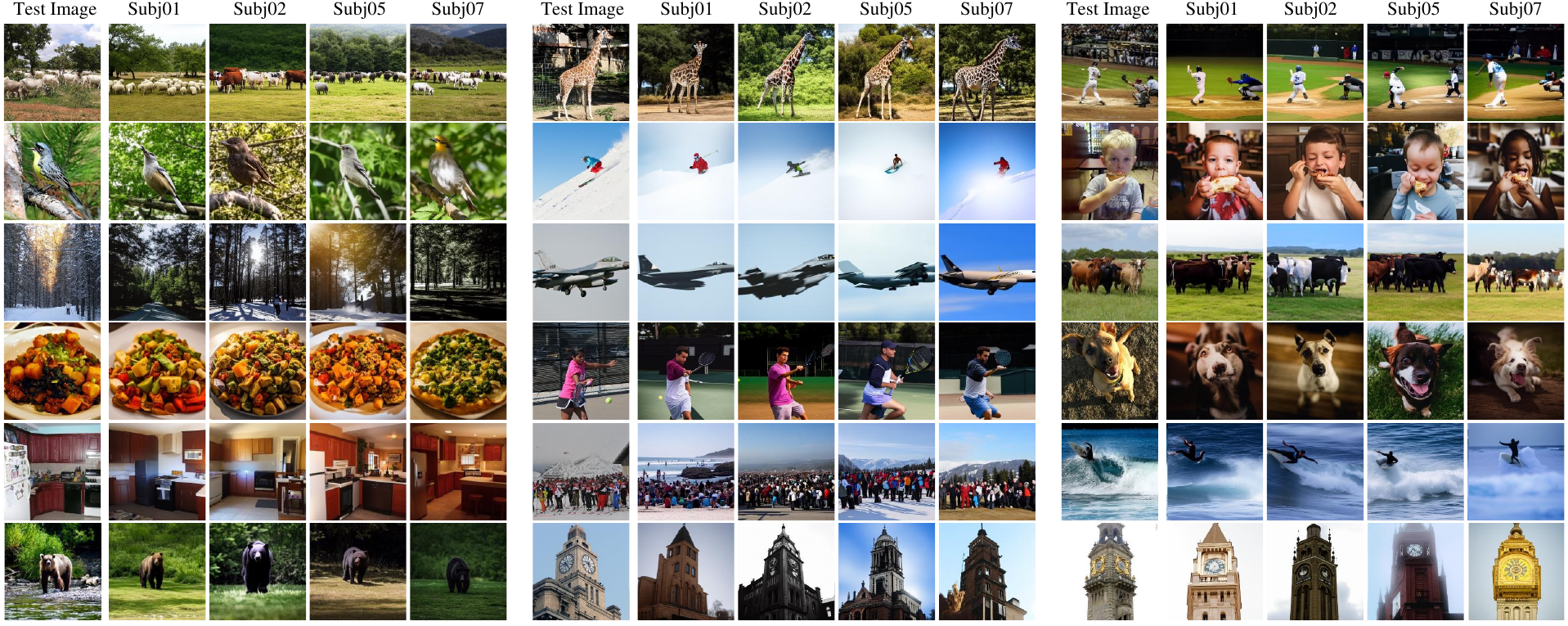}}
    \end{center}
   \caption{Reconstructed examples for 4 subjects}
    \label{fig:multi_subj}
\end{figure*}

A potential concern is that the observed improvement may be driven by the guidance aimed at optimizing image metrics, rather than by a comprehensive reconstruction enhancement. To address this concern, we have developed a set of brain correlation scores, which provide a unique measure of the relationship between reconstructed images and brain activity. These scores offer greater informativeness than image metrics. 
To illustrate this point, consider an extreme example, in which the guidance scale is elevated to 3,000,000. As depicted in Fig.\ref{fig:guidance_param}, the reconstructed images become distorted, a phenomenon that aligns with the model metrics. However, it is noteworthy that the brain correlation scores exhibit an increase in the early visual cortex, while the brain metrics remain unaffected by guidance functions. Consequently, we contend that the enhancement of \systemname\ may not be exclusively directed towards optimizing evaluation metrics, but rather, it is a genuine endeavor to enhance reconstruction quality. 

\begin{table}[!t]
    \centering
    \caption{Evaluated metrics for 4 subjects}
    \label{tab:multi_subj}
    \resizebox{1.\linewidth}{!}{
    \begin{tabular} {c|c|cccc}
        \toprule  
        \multicolumn{2}{c|}{Metrics} & subj01 & subj02 & subj05 & subj07 \\ 
        \midrule
        \multirow{4}{*}{Low-Level} & SSIM & 0.330 & 0.318 & 0.315 & 0.310 \\

        & Pixcorr & 0.378 & 0.309 & 0.261 & 0.247 \\

        & Alexnet(2) & 98.28\% & 96.58\% & 93.66\% & 92.78\% \\

        & Alexnet(5) & 99.43\% & 98.71\% & 97.87\% & 97.12\%\\
        \midrule
        \multirow{4}{*}{High-Level} & CLIP & 95.35\% & 94.95\% & 96.87\% & 94.46\%\\

        & Inception & 95.8\% & 94.66\% & 96.47\% & 94.55\%\\

        & EffNet & 0.642 & 0.670 & 0.658 & 0.683 \\

        & SwAV & 0.380 & 0.404 & 0.402 & 0.418 \\
        \midrule
        \multirow{6}{*}{Brain Corr.} & V1 & 0.407 & 0.386 & 0.331 & 0.303 \\
        & V2 & 0.374 & 0.335 & 0.315 & 0.282 \\
        & V3 & 0.352 & 0.335 & 0.293 & 0.262 \\
        & V4 & 0.320 & 0.335 & 0.284 & 0.245 \\
        & Higher Vis. & 0.346 & 0.358 & 0.393 & 0.287 \\
        & Whole Vis. & 0.365 & 0.367 & 0.386 & 0.295 \\
        \midrule
        \multirow{2}{*}{Retrieve} & Image & 99.96\% & 99.98\% & 98.48\% & 96.82\%\\
        & Brain & 99.92\% & 99.94\% & 95.32\% & 94.96\%\\
        \bottomrule
    \end{tabular}
    }
\end{table}

\subsubsection{Repeated Reconstruction}
We repeatedly generate five images with and without primary visual feature guidance for ground truth (GT) embeddings and fMRI embeddings, as illustrated in Fig.~\ref{fig:onetomany}. Reconstruction using GT embeddings represents the upper limit of performance and is expected to be consistent with the test images. However, reconstruction without guidance yields five different results, none of which are original. As analyzed in this article, the stable diffusion model is a generative model that produces diverse results, generating different images even with the same input. \systemname\ guides stable diffusion with detailed information, thus achieving an upper limit that conforms to the original image. When comparing fMRI embeddings with and without guidance, reconstruction with guidance exhibits clear detail fidelity (background, structure, color, texture) and consistency across the five repetitions.

\subsubsection{Reconstruction with Different Subjects}
The observed differences among subjects are not exclusively attributable to the differential representation of the fMRI decoder; they also imply subject-specific attention to distinct visual features in an image. For instance, as depicted in the bottom right of Fig.~\ref{fig:multi_subj}, subject 1 focuses on the overall perspective of the image, subjects 2 and 5 direct their attention to the building, and subject 7 focuses on the golden outline of the clock. 

\section{Discussion}
In this article, we propose a primary visual feature guidance for the stable diffusion model, with the objective of enhancing detail fidelity in fMRI-based visual stimulus reconstruction tasks. We analyze previous diffusion-based methods from a neuroscientific perspective and elucidate the reasons why the reconstructed images are semantically coherent yet detail-inaccurate. To address these limitations, we have developed \systemname\ , which incorporates additional primary visual features into the diffusion process, reflecting the bottom-up and top-down interactions within the visual cognitive mechanism, thereby improving detail fidelity. The proposed method involves two key modules: an fMRI decoder, which is trained to align fMRI data with the stable diffusion model, and a guidance method that integrates primary visual features into the diffusion process. The subsequent sections will provide a detailed discussion of each module.

\textbf{CLIP image embedding or text embedding:} The initial diffusion-based approach \cite{takagi2023high} employed ridge regression to decode fMRI data into CLIP text embeddings, leveraging the shared CLIP space to enhance the semantic coherence of the reconstructed images, surpassing the blurred results of GANs and VAEs. However, as this article underscores, text embeddings fall short in capturing fine-grained detail features, signifying avenues for improvement. MindEye \cite{scotti2023reconstructing} replaced CLIP text embeddings with CLIP image embeddings and utilized a large decoding model to achieve remarkable performance. We emphasize the significance of incorporating insights from the human visual system, encompassing both top-down and bottom-up processes. In this regard, text embeddings offer more pure high-level semantic features compared to image embeddings. However, we do not compromise on reconstruction details, which are enhanced through other forms guided by primary visual features. We believe that \systemname\  more closely aligns with the human visual system, thereby inspiring highly interpretable insights and further advancing brain decoding methods.

\begin{table}[!t]
    \centering
    \caption{Evaluated metrics for brain encoding guidance}
    \label{tab:multi_subj_boi}
    \resizebox{1.\linewidth}{!}{
    \begin{tabular} {c|c|cccc}
        \toprule  
        \multicolumn{2}{c|}{Metrics} & subj01 & subj02 & subj05 & subj07 \\ 
        \midrule
        \multirow{4}{*}{Low-Level} & SSIM & 0.334 & 0.319 & 0.310 & 0.305 \\

        & Pixcorr & 0.376 & 0.300 & 0.261 & 0.240 \\

        & Alexnet(2) & 98.18\% & 95.35\% & 91.47\% & 90.6\% \\

        & Alexnet(5) & 98.92\% & 97.82\% & 96.72\% & 95.66\%\\
        \midrule
        \multirow{4}{*}{High-Level} & CLIP & 95.07\% & 94.47\% & 96.35\% & 93.71\%\\

        & Inception & 95.8\% & 94.39\% & 96.38\% & 94.08\%\\

        & EffNet & 0.666 & 0.691 & 0.682 & 0.7 \\

        & SwAV & 0.402 & 0.426 & 0.424 & 0.436 \\
        \midrule
        \multirow{6}{*}{Brain Corr.} & V1 & 0.422 & 0.401 & 0.347 & 0.320 \\
        & V2 & 0.404 & 0.371 & 0.339 & 0.315 \\
        & V3 & 0.398 & 0.391 & 0.328 & 0.303 \\
        & V4 & 0.358 & 0.379 & 0.317 & 0.276 \\
        & Higher Vis. & 0.368 & 0.388 & 0.426 & 0.307 \\
        & Whole Vis. & 0.392 & 0.399 & 0.416 & 0.321 \\
        \bottomrule
    \end{tabular}
    }
\end{table}

\begin{figure*}[!t]
    \begin{center}
        \centerline{\includegraphics[width=1\linewidth]{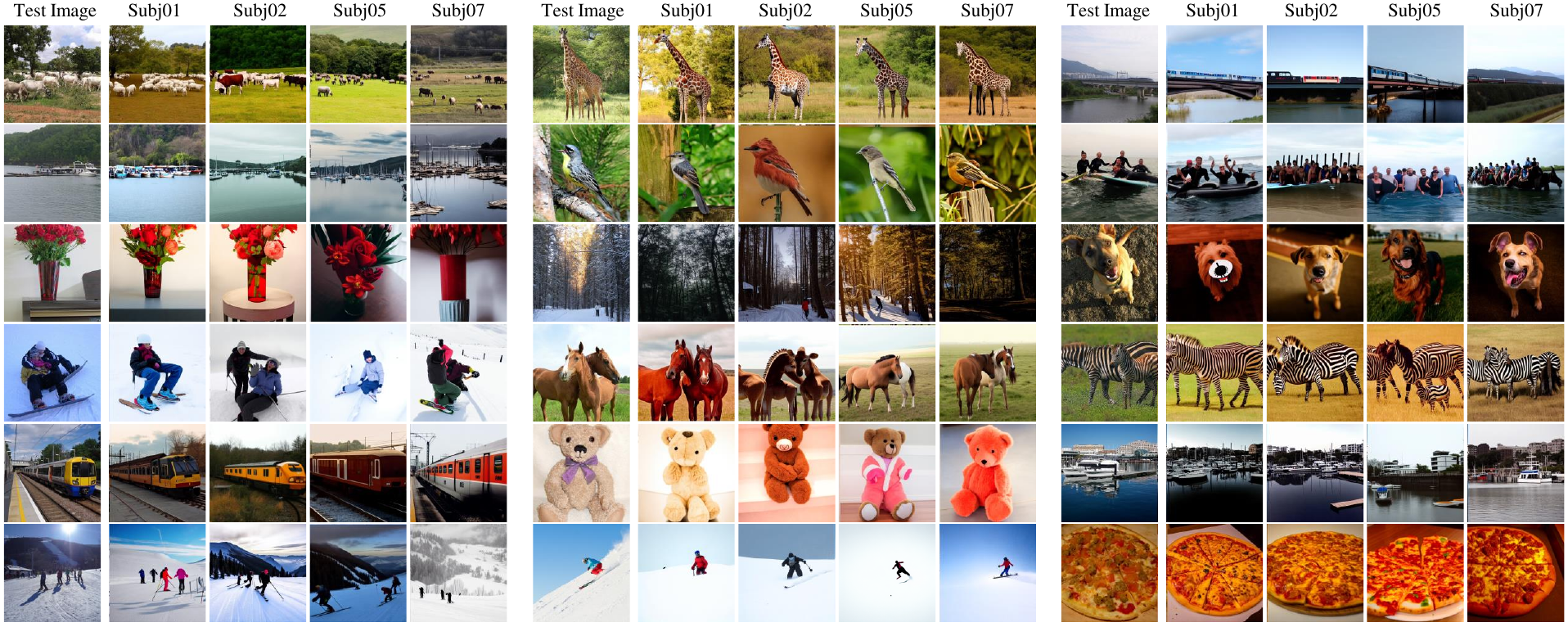}}
    \end{center}
   \caption{Reconstructed examples for brain encoding guidance}
    \label{fig:multi_subj_boi}
\end{figure*}

\textbf{Guidance function:} The proposed guidance method is universally applicable, facilitating a range of functions, including the brain encoding model (image to fMRI). It constitutes a visual reconstruction framework that integrates fMRI to image forward generation with an image to fMRI reverse guidance. We implemented this architecture using the brain encoding model (GNet~\cite{st2022brain}) as the guidance function, and the results are presented in Fig.~\ref{fig:multi_subj_boi} and Tab.~\ref{tab:multi_subj_boi}. It is imperative to acknowledge that the modification of the guidance function requires recalibration of the guidance hyperparameters. For this instance, the values of $eta$ and $gamma$ are set to 800 and 1.0, respectively.

\textbf{Image quality or detail fidelity:} We set two hyperparameters to manage the guidance strategy: the guidance scale, which influences the weight of details, and the guidance strength, which controls the number of guidance steps. Fig.~\ref{fig:guidance_param} and Tab.~\ref{tab:guidance_param} illustrate the qualitative and quantitative evaluations with different hyperparameters. The findings indicate that increasing the guidance scale improves low-level detail accuracy, but it also leads to a decline in overall image quality, as reflected in IS and FID metrics. Even with a staged guidance strategy, this reduction can be mitigated but not entirely eliminated. Thus, it is crucial to consider the trade-off: a larger guidance scale prioritizes detail fidelity, whereas a reduced guidance scale enhances image quality.

\textbf{Computational cost:} 
\textbf{1. Memory requirements:} \systemname\ can be trained and tested on a \texttt{NVIDIA GeForce RTX 3090 24GB} GPU. The hierarchical design of \systemname\ effectively separates semantic features from detail features, allowing for independent training of decoding models. Specifically, the decoder for semantic features (text embedding with dimensions of 77 × 768) has a total of 845M parameters for pre-training and 497M for subject-specific modeling, while the decoder for detail guidance features (feature layer of the CLIP image encoder with dimensions of 50 × 768) comprises 760M parameters for pre-training and 412M for subject-specific modeling. Both are lower than the 1412M/1064M parameters required for CLIP image embedding with dimensions of 257 × 768. It is noteworthy that while the total model size is substantial (1$\times$845M/497M + 6$\times$760M/412M = 5405M/2969M), each sub-model is relatively small and can be processed sequentially to reduce memory requirements.

\textbf{2. Time cost:} Training a decoding model on a \texttt{NVIDIA GeForce RTX 3090} GPU requires approximately 10 hours. Distributed Data Parallelism with multiple GPUs has been shown to accelerate model training. During the inference stage, we utilize Stable Diffusion v1.4 for image reconstruction, implementing 50 steps of DDIMSampler. Notably, \systemname\ does not result in a substantial increase in inference time, despite the face that the primary visual feature guidance necessitates more time due to the computation of gradients, as it is only employed for a limited number of steps. On a single 3090 GPU, the inference times with and without guidance are 5 seconds and 3 seconds, respectively, which is considerably lower than the optimization-based methods involving multiple diffusion processes.

\textbf{Limitation:} This paper demonstrates the potential of primary visual feature guidance for reconstructing human retinal stimulation. However, several challenges must be addressed for practical application:
\begin{enumerate}
    \item User-independent model: Most current studies focus on voxel activation in subject-specific visual cortex regions, limiting generalizability due to individual physiological differences. We need to explore generalized decoding models.
    \item fMRI hardware limitations: Practical applications are hindered by fMRI hardware constraints. Investigating retinal stimulation reconstruction using lightweight brain measurement technologies (e.g., EEG) would be more valuable.
    \item Inducing brain activation patterns: Brain activity from active imagination is a more powerful driver in the field of brain-computer interfaces than passive image stimulation.
\end{enumerate}

\section{Conclusion}
For the challenges of detail fidelity and consistency of repeated results in fMRI reconstruction, this paper presents a neuroscientific perspective on LDM-based methods. We attribute the presence of unfaithful details in LDM-based methods to the absence of a detail-driven bottom-up process. In light of these observations, we propose \systemname, a primary visual feature-guided LDM-based fMRI reconstruction method that incorporates detail cues to achieve detail fidelity without compromising semantic coherence. Furthermore, we introduce a novel guidance strategy for reconstruction tasks that utilizes hyperparameters $\kappa$ and $\eta$ to enhance the consistency of repeated results with original images. 

\medskip
{
\bibliography{ref}

\begin{thebibliography}{57}
\providecommand{\natexlab}[1]{#1}
\providecommand{\url}[1]{\texttt{#1}}
\expandafter\ifx\csname urlstyle\endcsname\relax
  \providecommand{\doi}[1]{doi: #1}\else
  \providecommand{\doi}{doi: \begingroup \urlstyle{rm}\Url}\fi

\bibitem[Wu et~al.(2020)Wu, Zhu, Wang, Zheng, and Chen]{wu2020encoding}
Hao Wu, Ziyu Zhu, Jiayi Wang, Nanning Zheng, and Badong Chen.
\newblock An encoding framework with brain inner state for natural image
  identification.
\newblock \emph{IEEE Transactions on Cognitive and Developmental Systems},
  13\penalty0 (3):\penalty0 453--464, 2020.

\bibitem[St-Yves and Naselaris(2018)]{st2018generative}
Ghislain St-Yves and Thomas Naselaris.
\newblock Generative adversarial networks conditioned on brain activity
  reconstruct seen images.
\newblock In \emph{2018 IEEE International Conference on Systems, Man, and
  Cybernetics (SMC)}, pages 1054--1061. IEEE, 2018.

\bibitem[Seeliger et~al.(2018)Seeliger, G{\"u}{\c{c}}l{\"u}, Ambrogioni,
  G{\"u}{\c{c}}l{\"u}t{\"u}rk, and van Gerven]{seeliger2018generative}
Katja Seeliger, Umut G{\"u}{\c{c}}l{\"u}, Luca Ambrogioni, Yagmur
  G{\"u}{\c{c}}l{\"u}t{\"u}rk, and Marcel~AJ van Gerven.
\newblock Generative adversarial networks for reconstructing natural images
  from brain activity.
\newblock \emph{NeuroImage}, 181:\penalty0 775--785, 2018.

\bibitem[Shen et~al.(2019)Shen, Dwivedi, Majima, Horikawa, and
  Kamitani]{shen2019end}
Guohua Shen, Kshitij Dwivedi, Kei Majima, Tomoyasu Horikawa, and Yukiyasu
  Kamitani.
\newblock End-to-end deep image reconstruction from human brain activity.
\newblock \emph{Frontiers in Computational Neuroscience}, 13:\penalty0 21,
  2019.

\bibitem[Lin et~al.(2019)Lin, Li, and Wang]{lin2019dcnn}
Yunfeng Lin, Jiangbei Li, and Hanjing Wang.
\newblock Dcnn-gan: Reconstructing realistic image from fmri.
\newblock In \emph{2019 16th International Conference on Machine Vision
  Applications (MVA)}, pages 1--6. IEEE, 2019.

\bibitem[Ren et~al.(2021)Ren, Li, Xue, Li, Yang, Jiao, and
  Gao]{ren2021reconstructing}
Ziqi Ren, Jie Li, Xuetong Xue, Xin Li, Fan Yang, Zhicheng Jiao, and Xinbo Gao.
\newblock Reconstructing seen image from brain activity by visually-guided
  cognitive representation and adversarial learning.
\newblock \emph{NeuroImage}, 228:\penalty0 117602, 2021.

\bibitem[Rombach et~al.(2022)Rombach, Blattmann, Lorenz, Esser, and
  Ommer]{rombach2022high}
Robin Rombach, Andreas Blattmann, Dominik Lorenz, Patrick Esser, and Bj{\"o}rn
  Ommer.
\newblock High-resolution image synthesis with latent diffusion models.
\newblock In \emph{Proceedings of the IEEE/CVF Conference on Computer Vision
  and Pattern Recognition}, pages 10684--10695, 2022.

\bibitem[Takagi and Nishimoto(2023{\natexlab{a}})]{takagi2023high}
Yu~Takagi and Shinji Nishimoto.
\newblock High-resolution image reconstruction with latent diffusion models
  from human brain activity.
\newblock In \emph{Proceedings of the IEEE/CVF Conference on Computer Vision
  and Pattern Recognition}, pages 14453--14463, 2023{\natexlab{a}}.

\bibitem[Ozcelik and VanRullen(2023)]{ozcelik2023natural}
Furkan Ozcelik and Rufin VanRullen.
\newblock Natural scene reconstruction from fmri signals using generative
  latent diffusion.
\newblock \emph{Scientific Reports}, 13\penalty0 (1):\penalty0 15666, 2023.

\bibitem[Lu et~al.(2023)Lu, Du, Zhou, Wang, and He]{lu2023minddiffuser}
Yizhuo Lu, Changde Du, Qiongyi Zhou, Dianpeng Wang, and Huiguang He.
\newblock Minddiffuser: Controlled image reconstruction from human brain
  activity with semantic and structural diffusion.
\newblock In \emph{Proceedings of the 31st ACM International Conference on
  Multimedia}, pages 5899--5908, 2023.

\bibitem[Scotti et~al.(2023)Scotti, Banerjee, Goode, Shabalin, Nguyen, Cohen,
  Dempster, Verlinde, Yundler, Weisberg, et~al.]{scotti2023reconstructing}
Paul~S Scotti, Atmadeep Banerjee, Jimmie Goode, Stepan Shabalin, Alex Nguyen,
  Ethan Cohen, Aidan~J Dempster, Nathalie Verlinde, Elad Yundler, David
  Weisberg, et~al.
\newblock Reconstructing the mind's eye: fmri-to-image with contrastive
  learning and diffusion priors.
\newblock In \emph{Proceedings of the 37th International Conference on Neural
  Information Processing Systems}, pages 24705--24728, 2023.

\bibitem[Ramesh et~al.(2022)Ramesh, Dhariwal, Nichol, Chu, and
  Chen]{ramesh2022hierarchical}
Aditya Ramesh, Prafulla Dhariwal, Alex Nichol, Casey Chu, and Mark Chen.
\newblock Hierarchical text-conditional image generation with clip latents.
\newblock \emph{ArXiv Preprint ArXiv:2204.06125}, 1\penalty0 (2):\penalty0 3,
  2022.

\bibitem[Allen et~al.(2022)Allen, St-Yves, Wu, Breedlove, Prince, Dowdle, Nau,
  Caron, Pestilli, Charest, et~al.]{allen2022massive}
Emily~J Allen, Ghislain St-Yves, Yihan Wu, Jesse~L Breedlove, Jacob~S Prince,
  Logan~T Dowdle, Matthias Nau, Brad Caron, Franco Pestilli, Ian Charest,
  et~al.
\newblock A massive 7t fmri dataset to bridge cognitive neuroscience and
  artificial intelligence.
\newblock \emph{Nature Neuroscience}, 25\penalty0 (1):\penalty0 116--126, 2022.

\bibitem[Kneeland et~al.(2023)Kneeland, Ojeda, St-Yves, and
  Naselaris]{kneeland2023brain}
Reese Kneeland, Jordyn Ojeda, Ghislain St-Yves, and Thomas Naselaris.
\newblock Brain-optimized inference improves reconstructions of fmri brain
  activity.
\newblock \emph{ArXiv Preprint ArXiv:2312.07705}, 2023.

\bibitem[Takagi and Nishimoto(2023{\natexlab{b}})]{takagi2023improving}
Yu~Takagi and Shinji Nishimoto.
\newblock Improving visual image reconstruction from human brain activity using
  latent diffusion models via multiple decoded inputs.
\newblock \emph{ArXiv Preprint ArXiv:2306.11536}, 2023{\natexlab{b}}.

\bibitem[Xia et~al.(2024)Xia, de~Charette, Oztireli, and Xue]{xia2024dream}
Weihao Xia, Raoul de~Charette, Cengiz Oztireli, and Jing-Hao Xue.
\newblock Dream: Visual decoding from reversing human visual system.
\newblock In \emph{Proceedings of the IEEE/CVF Winter Conference on
  Applications of Computer Vision}, pages 8226--8235, 2024.

\bibitem[Goodfellow et~al.(2014)Goodfellow, Pouget-Abadie, Mirza, Xu,
  Warde-Farley, Ozair, Courville, and Bengio]{goodfellow2014generative}
Ian Goodfellow, Jean Pouget-Abadie, Mehdi Mirza, Bing Xu, David Warde-Farley,
  Sherjil Ozair, Aaron Courville, and Yoshua Bengio.
\newblock Generative adversarial nets.
\newblock \emph{Advances in Neural Information Processing Systems}, 27, 2014.

\bibitem[Kingma and Welling(2013)]{kingma2013auto}
Diederik~P Kingma and Max Welling.
\newblock Auto-encoding variational bayes.
\newblock \emph{ArXiv Preprint ArXiv:1312.6114}, 2013.

\bibitem[Beliy et~al.(2019)Beliy, Gaziv, Hoogi, Strappini, Golan, and
  Irani]{beliy2019voxels}
Roman Beliy, Guy Gaziv, Assaf Hoogi, Francesca Strappini, Tal Golan, and Michal
  Irani.
\newblock From voxels to pixels and back: self-supervision in natural-image
  reconstruction from fmri.
\newblock In \emph{Proceedings of the 33rd International Conference on Neural
  Information Processing Systems}, pages 6517--6527, 2019.

\bibitem[Perarnau et~al.(2016)Perarnau, Van De~Weijer, Raducanu, and
  {\'A}lvarez]{perarnau2016invertible}
Guim Perarnau, Joost Van De~Weijer, Bogdan Raducanu, and Jose~M {\'A}lvarez.
\newblock Invertible conditional gans for image editing.
\newblock \emph{ArXiv Preprint ArXiv:1611.06355}, 2016.

\bibitem[Karras et~al.(2020{\natexlab{a}})Karras, Aittala, Hellsten, Laine,
  Lehtinen, and Aila]{karras2020training}
Tero Karras, Miika Aittala, Janne Hellsten, Samuli Laine, Jaakko Lehtinen, and
  Timo Aila.
\newblock Training generative adversarial networks with limited data.
\newblock \emph{Advances in Neural Information Processing Systems},
  33:\penalty0 12104--12114, 2020{\natexlab{a}}.

\bibitem[Karras et~al.(2020{\natexlab{b}})Karras, Laine, Aittala, Hellsten,
  Lehtinen, and Aila]{karras2020analyzing}
Tero Karras, Samuli Laine, Miika Aittala, Janne Hellsten, Jaakko Lehtinen, and
  Timo Aila.
\newblock Analyzing and improving the image quality of stylegan.
\newblock In \emph{Proceedings of the IEEE/CVF Conference on Computer Vision
  and Pattern Recognition}, pages 8110--8119, 2020{\natexlab{b}}.

\bibitem[Lin et~al.(2022)Lin, Sprague, and Singh]{lin2022mind}
Sikun Lin, Thomas Sprague, and Ambuj~K Singh.
\newblock Mind reader: Reconstructing complex images from brain activities.
\newblock \emph{Advances in Neural Information Processing Systems},
  35:\penalty0 29624--29636, 2022.

\bibitem[Ozcelik et~al.(2022)Ozcelik, Choksi, Mozafari, Reddy, and
  VanRullen]{ozcelik2022reconstruction}
Furkan Ozcelik, Bhavin Choksi, Milad Mozafari, Leila Reddy, and Rufin
  VanRullen.
\newblock Reconstruction of perceived images from fmri patterns and semantic
  brain exploration using instance-conditioned gans.
\newblock In \emph{2022 International Joint Conference on Neural Networks
  (IJCNN)}, pages 1--8. IEEE, 2022.

\bibitem[Gu et~al.(2024)Gu, Jamison, Kuceyeski, and Sabuncu]{gu2023decoding}
Zijin Gu, Keith Jamison, Amy Kuceyeski, and Mert~R Sabuncu.
\newblock Decoding natural image stimuli from fmri data with a surface-based
  convolutional network.
\newblock In \emph{Medical Imaging with Deep Learning}, pages 107--118. PMLR,
  2024.

\bibitem[Chen et~al.(2023)Chen, Qing, Xiang, Yue, and Zhou]{chen2023seeing}
Zijiao Chen, Jiaxin Qing, Tiange Xiang, Wan~Lin Yue, and Juan~Helen Zhou.
\newblock Seeing beyond the brain: Conditional diffusion model with sparse
  masked modeling for vision decoding.
\newblock In \emph{Proceedings of the IEEE/CVF Conference on Computer Vision
  and Pattern Recognition}, pages 22710--22720, 2023.

\bibitem[Scotti et~al.(2024)Scotti, Tripathy, Torrico, Kneeland, Chen, Narang,
  Santhirasegaran, Xu, Naselaris, Norman, and Abraham]{scotti2024mindeye2}
Paul~Steven Scotti, Mihir Tripathy, Cesar Torrico, Reese Kneeland, Tong Chen,
  Ashutosh Narang, Charan Santhirasegaran, Jonathan Xu, Thomas Naselaris,
  Kenneth~A. Norman, and Tanishq~Mathew Abraham.
\newblock Mindeye2: Shared-subject models enable f{MRI}-to-image with 1 hour of
  data.
\newblock In \emph{Proceedings of the 41st International Conference on Machine
  Learning}, pages 44038--44059, 2024.

\bibitem[Liu et~al.(2024)Liu, Ma, Zhu, Jing, and Zheng]{liu2024see}
Yulong Liu, Yongqiang Ma, Guibo Zhu, Haodong Jing, and Nanning Zheng.
\newblock See through their minds: Learning transferable neural representation
  from cross-subject fmri.
\newblock \emph{ArXiv Preprint ArXiv:2403.06361}, 2024.

\bibitem[Dhariwal and Nichol(2021)]{dhariwal2021diffusion}
Prafulla Dhariwal and Alex Nichol.
\newblock Diffusion models beat gans on image synthesis.
\newblock In \emph{Proceedings of the 35th International Conference on Neural
  Information Processing Systems}, pages 8780--8794, 2021.

\bibitem[Ho and Salimans(2022)]{ho2022classifier}
Jonathan Ho and Tim Salimans.
\newblock Classifier-free diffusion guidance.
\newblock \emph{ArXiv Preprint ArXiv:2207.12598}, 2022.

\bibitem[Zhang et~al.(2023)Zhang, Rao, and Agrawala]{zhang2023adding}
Lvmin Zhang, Anyi Rao, and Maneesh Agrawala.
\newblock Adding conditional control to text-to-image diffusion models.
\newblock In \emph{Proceedings of the IEEE/CVF International Conference on
  Computer Vision}, pages 3836--3847, 2023.

\bibitem[Mou et~al.(2024)Mou, Wang, Xie, Wu, Zhang, Qi, and Shan]{mou2023t2i}
Chong Mou, Xintao Wang, Liangbin Xie, Yanze Wu, Jian Zhang, Zhongang Qi, and
  Ying Shan.
\newblock T2i-adapter: Learning adapters to dig out more controllable ability
  for text-to-image diffusion models.
\newblock In \emph{Proceedings of the AAAI Conference on Artificial
  Intelligence}, volume~38, pages 4296--4304, 2024.

\bibitem[Zhao et~al.(2023)Zhao, Chen, Chen, Bao, Hao, Yuan, and
  Wong]{zhao2023uni}
Shihao Zhao, Dongdong Chen, Yen-Chun Chen, Jianmin Bao, Shaozhe Hao, Lu~Yuan,
  and Kwan-Yee~K Wong.
\newblock Uni-controlnet: all-in-one control to text-to-image diffusion models.
\newblock In \emph{Proceedings of the 37th International Conference on Neural
  Information Processing Systems}, pages 11127--11150, 2023.

\bibitem[Ju et~al.(2023)Ju, Zeng, Zhao, Wang, Zhang, and Xu]{ju2023humansd}
Xuan Ju, Ailing Zeng, Chenchen Zhao, Jianan Wang, Lei Zhang, and Qiang Xu.
\newblock Humansd: A native skeleton-guided diffusion model for human image
  generation.
\newblock In \emph{Proceedings of the IEEE/CVF International Conference on
  Computer Vision}, pages 15988--15998, 2023.

\bibitem[Voynov et~al.(2023)Voynov, Aberman, and Cohen-Or]{voynov2023sketch}
Andrey Voynov, Kfir Aberman, and Daniel Cohen-Or.
\newblock Sketch-guided text-to-image diffusion models.
\newblock In \emph{ACM SIGGRAPH 2023 Conference Proceedings}, pages 1--11,
  2023.

\bibitem[Zheng et~al.(2023)Zheng, Zhou, Li, Qi, Shan, and
  Li]{zheng2023layoutdiffusion}
Guangcong Zheng, Xianpan Zhou, Xuewei Li, Zhongang Qi, Ying Shan, and Xi~Li.
\newblock Layoutdiffusion: Controllable diffusion model for layout-to-image
  generation.
\newblock In \emph{Proceedings of the IEEE/CVF Conference on Computer Vision
  and Pattern Recognition}, pages 22490--22499, 2023.

\bibitem[Bansal et~al.(2023)Bansal, Chu, Schwarzschild, Sengupta, Goldblum,
  Geiping, and Goldstein]{bansal2023universal}
Arpit Bansal, Hong-Min Chu, Avi Schwarzschild, Soumyadip Sengupta, Micah
  Goldblum, Jonas Geiping, and Tom Goldstein.
\newblock Universal guidance for diffusion models.
\newblock In \emph{Proceedings of the IEEE/CVF Conference on Computer Vision
  and Pattern Recognition}, pages 843--852, 2023.

\bibitem[Pollen(1999)]{pollen1999neural}
Daniel~A Pollen.
\newblock On the neural correlates of visual perception.
\newblock \emph{Cerebral Cortex}, 9\penalty0 (1):\penalty0 4--19, 1999.

\bibitem[Gilbert and Sigman(2007)]{gilbert2007brain}
Charles~D Gilbert and Mariano Sigman.
\newblock Brain states: top-down influences in sensory processing.
\newblock \emph{Neuron}, 54\penalty0 (5):\penalty0 677--696, 2007.

\bibitem[Henderson et~al.(2023)Henderson, Tarr, and Wehbe]{henderson2023low}
Margaret~M Henderson, Michael~J Tarr, and Leila Wehbe.
\newblock Low-level tuning biases in higher visual cortex reflect the semantic
  informativeness of visual features.
\newblock \emph{Journal of Vision}, 23\penalty0 (4):\penalty0 8--8, 2023.

\bibitem[Rossion et~al.(2000)Rossion, Bodart, Pourtois, Thioux, Bol, Cosnard,
  Georges, Michel, and De~Volder]{rossion2000functional}
Bruno Rossion, Jean-Michel Bodart, Gilles Pourtois, Marc Thioux, Anne Bol, Guy
  Cosnard, Benoit Georges, Christian Michel, and Anne De~Volder.
\newblock Functional imaging of visual semantic processing in the human brain.
\newblock \emph{Cortex}, 36\penalty0 (4):\penalty0 579--591, 2000.

\bibitem[Kim et~al.(2020)Kim, Lee, Bae, and Yun]{kim2020mixco}
Sungnyun Kim, Gihun Lee, Sangmin Bae, and Se-Young Yun.
\newblock Mixco: Mix-up contrastive learning for visual representation.
\newblock \emph{ArXiv Preprint ArXiv:2010.06300}, 2020.

\bibitem[Song et~al.(2020)Song, Sohl-Dickstein, Kingma, Kumar, Ermon, and
  Poole]{song2021scorebased}
Yang Song, Jascha Sohl-Dickstein, Diederik~P Kingma, Abhishek Kumar, Stefano
  Ermon, and Ben Poole.
\newblock Score-based generative modeling through stochastic differential
  equations.
\newblock \emph{arXiv preprint arXiv:2011.13456}, 2020.

\bibitem[Efron(2011)]{efron2011tweedie}
Bradley Efron.
\newblock Tweedie’s formula and selection bias.
\newblock \emph{Journal of the American Statistical Association}, 106\penalty0
  (496):\penalty0 1602, 2011.

\bibitem[Kim and Ye(2021)]{kim2021noise2score}
Kwanyoung Kim and Jong~Chul Ye.
\newblock Noise2score: tweedie’s approach to self-supervised image denoising
  without clean images.
\newblock \emph{Advances in Neural Information Processing Systems},
  34:\penalty0 864--874, 2021.

\bibitem[Yang et~al.(2024)Yang, Gee, and Shi]{yang2023brain}
Huzheng Yang, James Gee, and Jianbo Shi.
\newblock Brain decodes deep nets.
\newblock In \emph{Proceedings of the IEEE/CVF Conference on Computer Vision
  and Pattern Recognition}, pages 23030--23040, 2024.

\bibitem[Wang et~al.(2023)Wang, Kay, Naselaris, Tarr, and
  Wehbe]{Wang2023NaturalLS}
Aria~Y Wang, Kendrick Kay, Thomas Naselaris, Michael~J Tarr, and Leila Wehbe.
\newblock Better models of human high-level visual cortex emerge from natural
  language supervision with a large and diverse dataset.
\newblock \emph{Nature Machine Intelligence}, 5\penalty0 (12):\penalty0
  1415--1426, 2023.

\bibitem[Lin et~al.(2014)Lin, Maire, Belongie, Hays, Perona, Ramanan,
  Doll{\'a}r, and Zitnick]{lin2014microsoft}
Tsung-Yi Lin, Michael Maire, Serge Belongie, James Hays, Pietro Perona, Deva
  Ramanan, Piotr Doll{\'a}r, and C~Lawrence Zitnick.
\newblock Microsoft coco: Common objects in context.
\newblock In \emph{European Conference on Computer Vision}, pages 740--755.
  Springer, 2014.

\bibitem[Wang et~al.(2004)Wang, Bovik, Sheikh, and Simoncelli]{wang2004image}
Zhou Wang, Alan~C Bovik, Hamid~R Sheikh, and Eero~P Simoncelli.
\newblock Image quality assessment: from error visibility to structural
  similarity.
\newblock \emph{IEEE Transactions on Image Processing}, 13\penalty0
  (4):\penalty0 600--612, 2004.

\bibitem[Krizhevsky et~al.(2012)Krizhevsky, Sutskever, and
  Hinton]{krizhevsky2012imagenet}
Alex Krizhevsky, Ilya Sutskever, and Geoffrey~E Hinton.
\newblock Imagenet classification with deep convolutional neural networks.
\newblock \emph{Advances in Neural Information Processing Systems}, 25, 2012.

\bibitem[Radford et~al.(2021)Radford, Kim, Hallacy, Ramesh, Goh, Agarwal,
  Sastry, Askell, Mishkin, Clark, et~al.]{radford2021learning}
Alec Radford, Jong~Wook Kim, Chris Hallacy, Aditya Ramesh, Gabriel Goh,
  Sandhini Agarwal, Girish Sastry, Amanda Askell, Pamela Mishkin, Jack Clark,
  et~al.
\newblock Learning transferable visual models from natural language
  supervision.
\newblock In \emph{International Conference on Machine Learning}, pages
  8748--8763. PMLR, 2021.

\bibitem[Szegedy et~al.(2016)Szegedy, Vanhoucke, Ioffe, Shlens, and
  Wojna]{szegedy2016rethinking}
Christian Szegedy, Vincent Vanhoucke, Sergey Ioffe, Jon Shlens, and Zbigniew
  Wojna.
\newblock Rethinking the inception architecture for computer vision.
\newblock In \emph{Proceedings of the IEEE Conference on Computer Vision and
  Pattern Recognition}, pages 2818--2826, 2016.

\bibitem[Tan and Le(2019)]{tan2019efficientnet}
Mingxing Tan and Quoc Le.
\newblock Efficientnet: Rethinking model scaling for convolutional neural
  networks.
\newblock In \emph{International Conference on Machine Learning}, pages
  6105--6114. PMLR, 2019.

\bibitem[Caron et~al.(2020)Caron, Misra, Mairal, Goyal, Bojanowski, and
  Joulin]{caron2020unsupervised}
Mathilde Caron, Ishan Misra, Julien Mairal, Priya Goyal, Piotr Bojanowski, and
  Armand Joulin.
\newblock Unsupervised learning of visual features by contrasting cluster
  assignments.
\newblock \emph{Advances in Neural Information Processing Systems},
  33:\penalty0 9912--9924, 2020.

\bibitem[St-Yves et~al.(2023)St-Yves, Allen, Wu, Kay, and
  Naselaris]{st2022brain}
Ghislain St-Yves, Emily~J Allen, Yihan Wu, Kendrick Kay, and Thomas Naselaris.
\newblock Brain-optimized deep neural network models of human visual areas
  learn non-hierarchical representations.
\newblock \emph{Nature communications}, 14\penalty0 (1):\penalty0 3329, 2023.

\bibitem[Salimans et~al.(2016)Salimans, Goodfellow, Zaremba, Cheung, Radford,
  and Chen]{salimans2016improved}
Tim Salimans, Ian Goodfellow, Wojciech Zaremba, Vicki Cheung, Alec Radford, and
  Xi~Chen.
\newblock Improved techniques for training gans.
\newblock In \emph{Proceedings of the 30th International Conference on Neural
  Information Processing Systems}, pages 2234--2242, 2016.

\bibitem[Seitzer(2020)]{Seitzer2020FID}
Maximilian Seitzer.
\newblock {pytorch-fid: FID Score for PyTorch}.
\newblock \url{https://github.com/mseitzer/pytorch-fid}, August 2020.
\newblock Version 0.3.0.

\end{thebibliography}
\bibliographystyle{unsrtnat}
}

\end{document}